\RequirePackage{amsmath}
\PassOptionsToPackage{caption=false}{subfig}
\documentclass{article}
\pdfoutput=1 %

\usepackage[final]{corl_2020} %

\usepackage{amsmath,amssymb} %
\usepackage{graphicx}
\usepackage{wrapfig}
\usepackage{comment}
\usepackage{multirow}

\usepackage{booktabs}
\usepackage{adjustbox}
\usepackage{sidecap}
\usepackage{bm}
\usepackage{svg}
\usepackage[caption=false]{subfig}
\usepackage{setspace}
\usepackage{sidecap}

\def\ShowAppendix{1}  %

\newcommand\best[1]{{\textbf{#1}}}
\newcommand\second[1]{{\underline{#1}}}

\makeatletter
\newcommand{\printfnsymbol}[1]{%
  \textsuperscript{\@fnsymbol{#1}}%
}
\makeatother

\usepackage{xspace}
\makeatletter
\DeclareRobustCommand\onedot{\futurelet\@let@token\@onedot}
\def\@onedot{\ifx\@let@token.\else.\null\fi\xspace}

\def\ie{\emph{i.e}\onedot}

\def\etal{\emph{et al}\onedot}
\makeatother

\title{Range Conditioned Dilated Convolutions for Scale Invariant 3D Object Detection}

\author{
  Alex Bewley~\thanks{Indicates equal contribution. Correspondence: \texttt{\{bewley, peis\}@google.com}}\\
  Google Research\\
  \And
  Pei Sun \printfnsymbol{1} \\
  Waymo LLC \\
  \AND
  Thomas Mensink \\
  Google Research\\
  \And
  Dragomir Anguelov \\
  Waymo LLC \\
  \And
  Cristian Sminchisescu \\
  Google Research\\
}

\begin{document}
\maketitle

\setcounter{footnote}{0} 
\begin{abstract}

This paper presents a novel 3D object detection framework that processes LiDAR data directly on its native representation: \textit{range images}. Benefiting from the compactness of range images, 2D convolutions can efficiently process dense LiDAR data of a scene. To overcome scale sensitivity in this perspective view, a novel range-conditioned dilation (RCD) layer is proposed to dynamically adjust a continuous dilation rate as a function of the measured range. Furthermore, localized soft range gating combined with a 3D box-refinement stage improves robustness in occluded areas, and produces overall more accurate bounding box predictions. On the public large-scale Waymo Open Dataset, our method sets a new baseline for range-based 3D detection, outperforming multiview and voxel-based methods over all ranges with unparalleled performance at long range detection.
\end{abstract}

\begin{figure}[h]
    \centering
    \includegraphics[width=\textwidth]{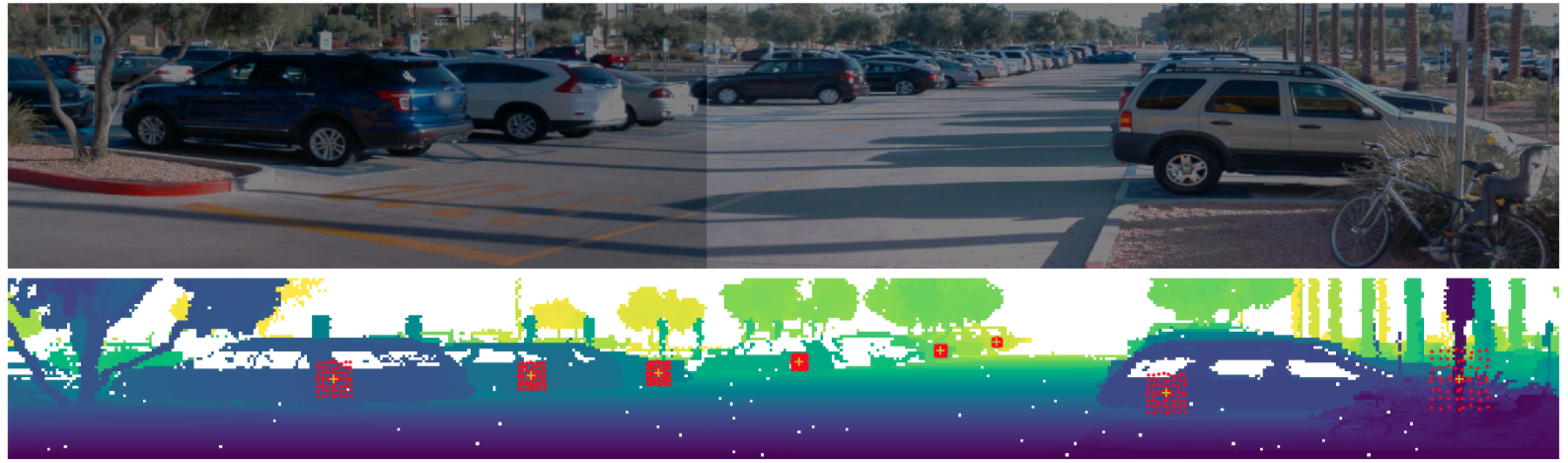}
	\caption{\textbf{Top:} Reference RGB images of a scene. \emph{For illustration purposes only.}  \textbf{Bottom:} Range image showing the dynamic sampling of our proposed RCD layer at selected positions. Here the measured range at the yellow \textcolor{orange}{$+$} is used to govern a scale of the local receptive field towards a geometrically consistent sample density at any range. 
	}
	\label{fig:short}
\end{figure}

\section{Introduction}
One of the most exciting opportunities at the intersection of robotics and machine learning is autonomous driving, where the detection of objects is of critical importance. Catalyzed by the proliferation of high-density LiDAR sensors, algorithms for 3D detection from pointcloud data have gained significant attention in recent years. %
Several meta-architectures are used for addressing 3D object detection. Firstly, voxelization methods~\cite{engelcke2017vote3deep,wang2015voting,zhou2018voxelnet} typically bin sparse Cartesian coordinates into discrete voxels and later process them with subsequent 3D convolutions. While such approaches perform well in practice, they are impeded by the memory and computational demands of 3D convolutions in large scenes. A related meta-architecture is projecting the sparse points into a birds-eye view (BEV) \cite{Chen_2017_CVPR,lang2019pointpillars} in order to reduce the scene back to a 2D space while maintaining scale invariance at the price of lost information through quantization. Other meta-architectures are built upon the PointNet framework \cite{qi2017pointnet,qi2017pointnet++} but suffer from issues around point sparsity at long range and often require inefficient custom operations for defining point neighborhoods~\cite{Engelmann20ICRA}.

This paper focuses on an a less explored alternative (in the context of 3D detection) using range images that exploits the intrinsic 2.5D manifold structure \cite{bewley2013advantages,wu2018squeezeseg} of raw 3D point data in its native spherical coordinate form, enabling efficient and direct 3D object detection. 
This \emph{range image} representation is characterised as 3D Cartesian points projected onto unique pixels in a 2D spherical image where range is encoded as pixel values and their row and column indices correspond to inclination and azimuth angles respectively.
Operating on range images enjoys the benefits of applying mature 2D convolutional architectures and is naturally efficient due to the  intrinsically compact representation \cite{meyer2019lasernet}. 
Crucially, it does not suffer from the issue of sparsity at long range.
However, known challenges for learning such as scale variation and occlusion need further consideration. This paper addresses these issues by proposing a novel convolutional layer with a scale aware dilation rate for efficient reuse of filter weights at different scales. This directly leverages the measure distance in range images to compensate for the corresponding scale change. Combined with soft range-based gating, both scale, and occlusion are appropriately handled within this framework.
Occlusion is further addressed through the use of a second stage local box refinement module.

In this work, we present an efficient range image-based two-stage 3D object detector with the following key contributions.
Firstly, a novel range conditioned dilated (RCD) convolutional operator is introduced that is capable of dynamically adjusting the local receptive field to provide a consistent scale relative to the convolutional kernel at any distance (see Figure \ref{fig:short}).
Second, a region convolutional neural network (RCNN) based second stage network is investigated in the context of range image-based 3D object detection;
Finally, a new baseline is set for range image-based 3D object detection on a public dataset~\cite{sun2019scalability}. 
The introduced RCD based model performs especially well at long ranges (large distances), where voxel and sparse-point cloud based approaches suffer from point sparsity issues. Therefore, we believe this is the first work to combine a range image-based network with a RCNN second stage for 3D object detection.

\section{Related Work}

\subsection{3D LiDAR Detection}
While many works combine color images with LiDAR data \cite{qi2018frustum,ku2018joint,wang2019frustum}, here we restrict our review to works that only process 3D LiDAR data.

\textbf{Rasterized and Voxel methods.} A popular way to do 3D object detection is to first project the points to birds-eye view (BEV) and constructs a 2D multi-channel image. The image is then processed by 2D CNN to get either BEV or 3D boxes. The transformation process is usually hand-crafted, some selected works MV3D~\cite{Chen_2017_CVPR}, PIXOR \cite{yang2018pixor}, Complex YOLO ~\cite{Simony_2018_ECCV_Workshops}. VoxelNet~\cite{zhou2018voxelnet} divides the point cloud into a 3D voxel grid and uses a PointNet-like network~\cite{qi2017pointnet} to learn an embedding of the points inside each voxel. PointPillars~\cite{lang2019pointpillars}, a compute efficient method, that divides the point cloud into 3D pillars and then extracts features similar as VoxelNet~\cite{zhou2018voxelnet}, is the most prolific detector in this category and serves as a comparison in our experiments. 

\textbf{Point based methods.} Another paradigm of methods are point based detection. It processes the raw point cloud with point cloud feature extraction methods like PointNet++~\cite{qi2017pointnet++}, Sparse Convolution~\cite{graham2017submanifold}, and then regresses 3D boxes in either downsampled BEV view or 3D point view directly. Some representative works are PointRCNN~\cite{shi2019pointrcnn}, PVRCNN~\cite{shi2019pv}, STD~\cite{yang2019std} or SA-SSD~\cite{he2020sassd} which uses pointwise supervision for training a voxel based backbone.
Our method is benchmarked against PVRCNN which is currently the top detector on the KITTI dataset and also benchmarked on Waymo Open Dataset~\cite{sun2019scalability}.

\textbf{Range image based methods.} 
The range image is compact and does not suffer from sparsity related issues which is the main challenge when developing 3D algorithms.
This representation is under-explored because a) range image based detectors require more data to train~\cite{meyer2019lasernet}. b) generating high quality range images is non-trivial without knowing raw sensor information such as laser scan pattern, relative position at each laser shot. Both of these are addressed by the Waymo Open Dataset~\cite{sun2019scalability}. The primary representative work is LaserNet~\cite{meyer2019lasernet} which benchmarks on a private dataset. Range image based detection algorithms need to deal with scale variance (near range objects are larger) and occlusions. %

\subsection{Adaptive and Dilated Convolutions}
The seminal work of Jaderberg \etal on transforming the input signal using spatial transformers~\cite{jaderberg2015spatial} has led to several subsequent methods.
For example, for dilated convolutions, the dilation rate could be learned per filter and layer~\cite{he2017learning}, or related to the size of the 3D environment using RGBD as input~\cite{chen20193d}.
Wang \etal~\cite{wang2018depth} use depth as a constant weighting factor of the influence of neighboring pixels in convolutional layers and max-pooling layers.
More generically, dynamic functions could be learned to generate the weights of the convolutional layer conditioned on its input values~\cite{dai2017deformable,strub2018visual,zhu2019deformable,ding2020learning}.
Ding \etal~\cite{ding2020learning} propose a dynamic filter framework which simulates a dilated convolution with integer shifts to the feature maps, these are then combined using weights from RGB image features. Here a fixed set of dilations are chosen a priori similar the atrous spatial pyramid pooling (ASPP) in DeepLab~\cite{chen2017deeplab}.
In contrast, our method takes advantage of the available range observations to rescale the dilation rate enabling the reuse of kernel weights across multiple scales. This can be viewed as an extension of the 1D distance based input filtering of Beyer \etal \cite{BeyerHermans2016RAL} to 3D pointclouds represented as a range image.
Furthermore, our sampler is continuous and adapts through the course of training mitigating the need to select a set of fixed dilation rates.

\section{Range Condition to Learn Scale Invariant Features}

This section describes the proposed range conditioned dilated convolutional block. This block can be placed anywhere in a larger network where a \emph{range image} is available. %

\subsection{Range Conditioned Dilated Convolution}
\begin{figure}[t]
	\begin{center}
		\includegraphics[width=0.95\textwidth]{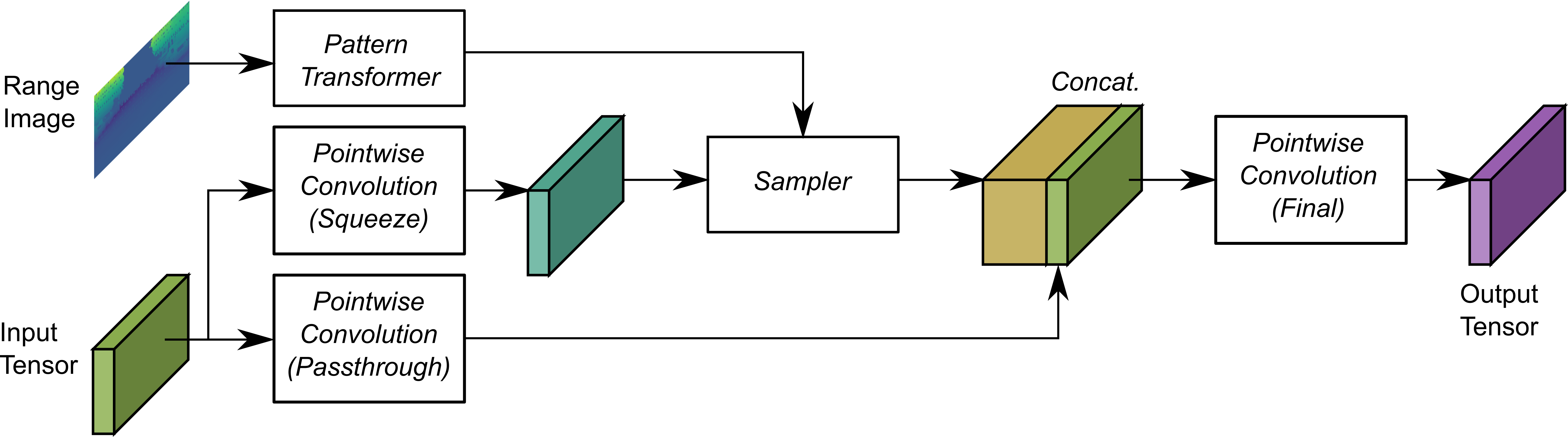}
	\end{center}
	\caption{An overview of the range-conditioned dilation block detailing the inputs and outputs the various modules (illustrated with white boxes). The sampler is solely responsible for the spatial processing of the input tensor where the receptive field is driven by the input range image.
	}
	\label{fig:rcd_block_overview}
\end{figure}

The RCD block accounts for the scale variation when observing objects at different distances by dynamically adjusting the spatial extent of a convolution via its dilatation rate.
The dilation is scaled using a \emph{continuous} function of the measure range requiring the replacement of discrete pixel index lookup (used in regular kernels) with a form of spatial sampling that is simultaneously sparse, local and conditioned on the observed range. 
As the dilation rate is continuous as opposed to integer-valued, a specialization of 
the spatial transformer~\cite{jaderberg2015spatial} is adapted to adjust the spatial scale of the sampling pattern. This sampling is applied densely to every pixel to create a form of deformable convolution~\cite{chen2017deeplab,yu2015multi} that is conditioned on the input range.
This section describes the RCD applied to 2D range-images relevant for this paper, however without loss of generality the equivalent operations can be applied to other dimensional inputs.
Figure \ref{fig:rcd_block_overview} provides a high-level overview of the RCD block with the individual components detailed below.

\begin{wrapfigure}{RT}{0.5\textwidth}
    \centering
    \includegraphics[width=0.5\textwidth]{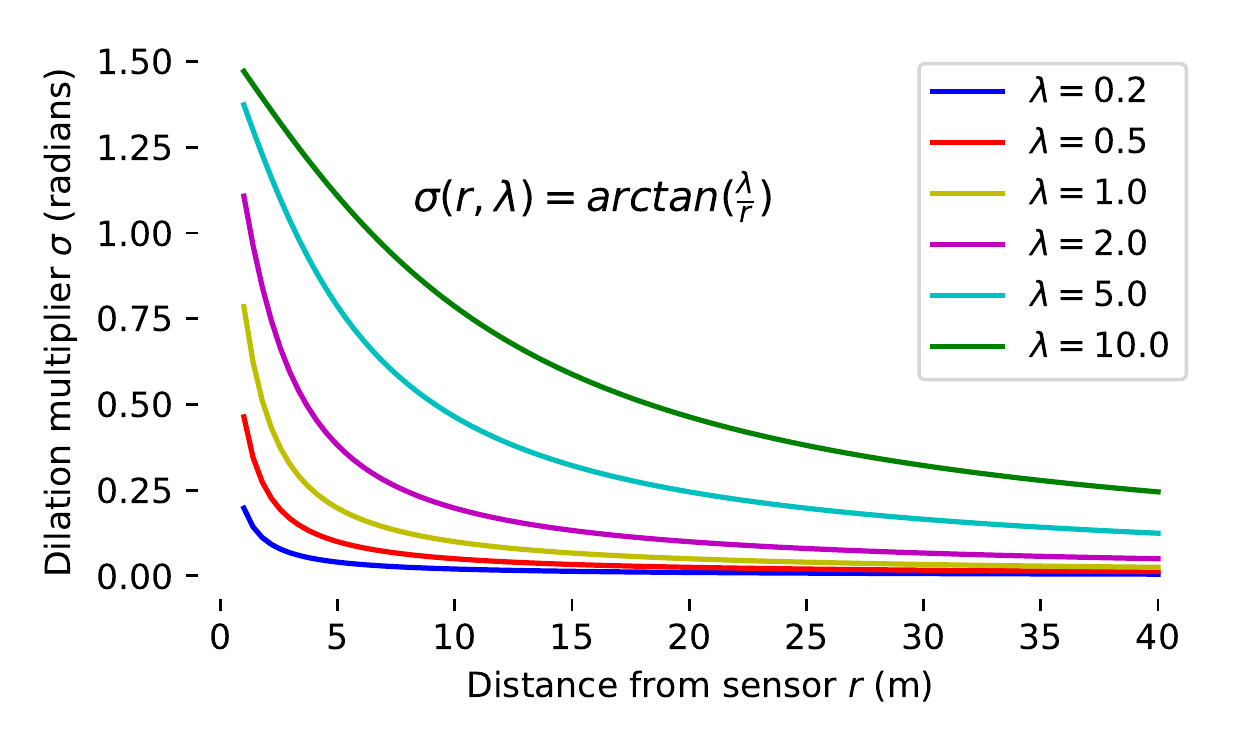}
    \caption{\label{fig:dynamic_dilation}The rate of the convolutional filter dilation as a function of both the distance and the nominal object size $\lambda$ in meters. Note multiplier is in units of radians and should be adjusted to pixels using the appropriate LiDAR angular resolution.}
\end{wrapfigure}

\textbf{Pattern Transformer:} takes in a range-image $R \in \mathbb{R}^{H \times W}$ with height $H$ and width $W$. Internally parameters in the form of $N$ sparse 2D points $\mathcal{G} \in \mathbb{R}^{N \times 2}$ are maintained to represent the relative spatial sampling pattern for the RCD convolutional kernel. The sampling pattern $\mathcal{G}$ is a set of learnable parameters which is initialized to be a uniform grid with mean $(0,0)$. This relative sampling pattern $\mathcal{G}$ is shared for all pixel locations and then individually transformed using the input range image as follows:
\begin{equation}
    \mathcal{S} = \sigma(R, \lambda) \cdot \mathcal{G} + \mathcal{P},
    \label{eq:transform_function}
\end{equation}
where (via broadcasting\footnote{Singular tensor shapes are omitted for brevity.}) $\mathcal{S} \in \mathbb{R}^{H \times W \times N \times 2}$, $\mathcal{P} \in \mathbb{R}^{H \times W \times 2}$ represents pixel coordinates and $\sigma(R, \lambda)$
is the following trigonometric function applied to all range values $r_i = R(i), i \in \mathcal{P}$:
\begin{equation}
    \sigma(r_i, \lambda) = \arctan(\lambda/r_i),
    \label{eq:scale_function}
\end{equation}
where $\lambda \in \mathbb{R}^+$ is a learnable scalar parameter representing the nominal width of the cross-sectional area covered by the receptive field at any range $r_i$. 
The dilation multiplier $\sigma(r_i, \lambda)$ is visualized for several values of  $\lambda$ in Figure \ref{fig:dynamic_dilation}.

\textbf{Sampler:} %
At the core of the RCD block is a sampler that is responsible for gathering spatial information from its input analogous to the dilated kernel sampling, ready for an inner product with kernel weights as with conventional convolutions.
Given input feature tensor $X_{s} \in \mathbb{R}^{H \times W \times C}$, input sampling is performed $N$ times for all pixel locations to produce a resampled tensor $\hat{X}_s \in \mathbb{R}^{H \times W \times N \times C}$, where $C$ is the dimensionality of pixelwise features in $X_s$. 
While any sampling kernel can be applied, the bilinear sampling is chosen for its efficiency~\cite{dai2017deformable} and its spatial partial derivatives~\cite{jaderberg2015spatial} allow for loss gradients to flow back to the size and spatial parameters $\lambda$ and $\mathcal{G}$.

The coordinates contained in $\mathcal{S}$ may include samples extending beyond the width or height of the input range image. 
Samples extending above and below image boundaries are clamped to the first and last row respectively. 
For samples beyond the left and right boundaries, horizontal angular wrapping is performed, taking advantage of the range image\'s $360^\circ$ view.

In parallel to scaling the perceptive field, the range is also used to weight the magnitude of sampled features. This mechanism serves as a form of soft range gating (SRG) to prevent down-stream training difficulties caused by distractors from near-occlusion where neighboring pixels have a substantial difference in range.
The weighting for SRG follows a Gaussian distribution, where for a given pixel location $i$ with corresponding sample location $j \in \mathcal{S}(i)$, a feature masking weight: %
\begin{equation}
    \bar{\bm x}_{ij} = {\bm \hat{x}}_{ij} \ \mathcal{N}(\hat{r}_j; r_i, \gamma),
\end{equation}
where $\mathcal{N}$ denotes the Gaussian probability density function, evaluated at the range of the integer spatial location $r_i$, with mean $\hat{r}_j = R(j)$ being the bilinearly interpolated range value at the sample location $j$.
The variance $\gamma$ is a learnable parameter that controls the length-scale of the soft range gate (initialized to 1 meter). 
This decreases the importance of distant points, which are likely from a different object to the point at the center of the convolution.

\textbf{Pointwise Convolutions:} Also known as $1 \times 1$ convolutions~\cite{lin2013network} are used in multiple parts of the RCD block. See Figure \ref{fig:rcd_block_overview} for their position within the RCD block. Firstly, two pointwise convolutions (PConv) partition the input tensor for the sampler input and concatenation with its output. This arrangement is inspired by the two stream hypothesis~\cite{goodale1992separate} where the sampler with its input PConv and the pass-through PConv are responsible for spatial and recognition processes respectively. On a practical note, the PConv feeding into the spatial sampler projects the input into lower dimensional features, significantly reducing the computation and memory requirements (in all RCD experiments $C=3$). 
The output of the sampler is reshaped to $\mathbb{R}^{H \times W \times NC}$ and then concatenated along the last dimension of the result of the pass-through PConv. 
The final PConv completes the convolutional operation as the input contains the dilated resampling of the input and is reshaped to pack the spatial samples into the channel dimension. The weights in this final PConv are essentially the weights of the RCD convolution making the final output equal to:
\begin{equation}
    \mathbf{h}_i = f(W^s_f \mathbf{x}^s_i + W^p_f \mathbf{x}^p_i),
\end{equation}
where $\mathbf{x}^s_i,\mathbf{x}^p_i$ are the flattened spatially sampled features and passthrough feature respectively at pixel location $i$ and $f(\cdot)$ represents Layer Normalization~\cite{clevert2015fast} followed by the a non-linear exponential linear unit \cite{ba2016layer}.

\section{Two Stage Detection Network}

\begin{figure*}[t!]
    \centering
    \includegraphics[width=\columnwidth]{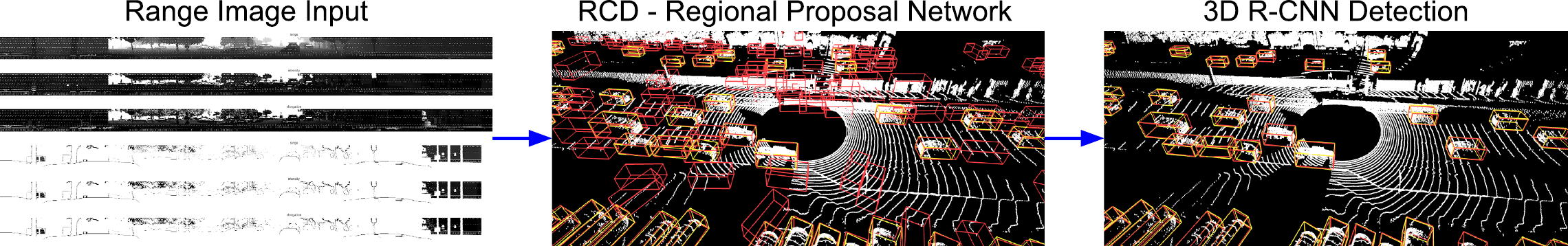}
    \caption{Two stage design: the input range images (size $64\times2650\times8$) are first provided to the RCD-RPN which generates a 3D box proposal corresponding to each point. High scoring proposals (red boxes) are sent to the second 3D RCNN stage for refinement.
    }
    \label{fig:sensor_layout}
    \label{fig:overview_all}
\end{figure*}

The RCD block is embedded within the region proposal network (RPN) stage of a two stage detection network, see Figure \ref{fig:overview_all}.
The first stage is an adaption of the LaserNet~\cite{meyer2019lasernet} architecture, with RCD blocks, containing both a foreground classification and box regressor head.
The second stage refines high-scoring box proposals similar in spirit to~\cite{shi2019pointrcnn, shi2019part}. %

\subsection{First stage: RCD-RPN}
The first stage is a fully convolutional deep layer aggregation network architecture~\cite{yu2018deep, meyer2019lasernet}, 
with RCD blocks replacing the initial convolutional layers at multiple scales (see supplementary).
To achieve a larger receptive field and faster compute, we downsample more aggressively along the horizontal axis only with a $[1, \cdot ]$ pooling kernel.
This befits the typical structure of range images with few rows and many columns.
The RCD-RPN network yields four outputs per point: a foreground classification score, two additional scores representing the probability its top and left neighboring pixels lay on the same object, and a predicted 3D box represented by a 7D vector ($x, y, z, h, w, l, \theta$). 

The focal loss~\cite{lin2017focal}: $L_f(p_i) = -\alpha_i (1-p_i)^\gamma \log(p_i)$ is employed, where $p_i = s_i$ if $i$ is positive, and $p_i = 1-s_i$ otherwise, to handle class imbalance (using $\alpha_i$) and focus on difficult boxes (using focus parameter $\gamma=2$), for the three classification scores:
\begin{equation}
L_{\textrm{f}} = \frac{1}{|\mathcal{P}|} \sum_i{[L_f(p_i) + L_{f}(p_i^{\textrm{t}}) + L_{f}(p_i^{\textrm{l}})]},
\label{eq:focal}
\end{equation}
where $|\mathcal{P}|$ is the number of valid points in the range image, and $p_i, p_i^{\textrm{t}}, p_i^{\textrm{l}}$ are the scores for foreground class and top and left similarity classification respectively.

For regressing towards the 3D box parameters, the bin-loss ($L_{bin}$)~\cite{shi2019pointrcnn} divides the distance to the center point ($x,y,z$) and the heading of the ground truth box into bins and performs bin classification first, followed by a regression within each bin (see~\cite{shi2019pointrcnn} for more details). It has been shown that the bin-loss converges faster and achieves higher recall than regression based loss.
For each point $i$ in the range image, we use its predicted 3D box $b_i$ as follows:
\begin{equation}
L_{\textrm{b}} = \frac{1}{N} \sum_i \tfrac{1}{n_{i}} L_{bin}(b_i, b_j),
\end{equation}
\noindent where $N$ is the number of ground-truth boxes, we sum over all pixels in the range image, $n_i$ is the number of points in the target ground-truth box $b_j$ that contains point $i$. 

\subsection{Second stage: 3D RCNN}
The second stage refines the initial box proposals, by using an RCNN stage similar to~\cite{ren2015faster,shi2019pointrcnn,shi2019part}.
In general a two stage design greatly improves box prediction accuracy and further mitigates the occlusion effect. 
For each box, the raw points are extracted and the per point feature embeddings, predicted box parameters and semantic classification scores are reused from RCD-RPN. 
Each box is transformed into the canonical box frame~\cite{shi2019pointrcnn} and divided into a a fixed 3D grid ($12\times8\times6$).
All the features in each grid cell are pooled to yield a single feature per grid cell. For the semantic features (points, classification score and RCD-RPN box parameters) average pooling is used, while the feature embeddings are max pooled.
Then a 3D convolution layer, followed by downsampling and a fully connected layer generates the final box parameters and classification score. 

During training, it is important to sub-sample boxes from the first stage to have an efficient training pipeline.  
First, the range image is divided into a top and bottom half, and per column the highest scoring box-proposal is kept. 
This coarse partitioning maintains spatial diversity for training the second stage.
The remaining proposals are then further reduced to 50 positive (with  intersection over union $IoU \geq .5$) and 50 negative ($IoU < .5$) proposals. During inference, the top 400 boxes are kept after running non-maximum-suppression on the RCD-RPN box proposals.

The second stage 3D RCNN also contains classification loss which is averaged over all box proposals ($M=100$ during training) selected from RCD-RPN:
\begin{equation}
L_{\textrm{cls}} = \frac{1}{M} \sum_k L_{ce}(s_k, y_k),
\end{equation}
where $s_k$ denotes the refined classification score, $y_k$ the ground-truth class for box $k$, and $L_{ce}(\cdot)$ the cross-entropy loss. 
For the refined bounding box parameters $b_k$ a combination of losses is used, including a residual loss similar to~\cite{ren2015faster} for box centers ($x,y,z$) and box dimensions ($w, h, l$), while heading ($\theta$) regression uses the bin loss~\cite{shi2019pointrcnn}. 
These losses are averaged over all proposals:
\begin{equation}
L_{\textrm{reg}} = \frac{1}{M} \sum_k L_{\textrm{box}}(b_j, b_k),
\label{eq:reg}
\end{equation}
where the box regression loss $L_{\textrm{box}}$ is a sum of smooth $L1$ loss for center positions and normalized length, width and height prediction and a bin loss for heading prediction.

\subsection{Joint Training: }
Both the RPN and RCNN networks are trained jointly with the final objective as:
\begin{equation}
L = \underbrace{L_{\textrm{f}} + L_{\textrm{b}}}_{\textrm{RCD-RPN}} + \underbrace{L_{\textrm{cls}} + L_{\textrm{reg}}}_{\textrm{3D RCNN}}.
\end{equation}

\section{Experiments}

We primarily benchmark on the Waymo Open Dataset (WOD)~\cite{sun2019scalability} as it released its raw data in range image format while other datasets such as KITTI \cite{geiger2013vision} or nuScenes \cite{nuscenes2019} provide pointclouds. 
Converting a pointcloud back into a range image requires known laser angles and accurate pointwise timing to offset for the relative vehicle pose when in motion. Simply projecting Cartesian points to their spherical counterparts~\cite{milioto2019rangenet} results in either significant pixel collisions or many holes depending on the choice of range-image resolution. For the KITTI LiDAR with 64 individual lasers, their unique vertical and inclination offsets are recovered with the Hough-transform and points are grouped to form the rows in the range image. Table \ref{tab:kitti_test} shows the BEV results on the KITTI detection testset.
See Section \ref{sec:kitti_results} in Appendix for more details.
As reported in \cite{meyer2019lasernet}, we also observe that small datasets are prone to overfitting for range image detectors and resume our evaluation using the larger WOD dataset.

\begin{SCtable}
	\adjustbox{width=.45\textwidth}{
        \begin{tabular}{lccc}
        \toprule
        {\bf Method} & {\bf Easy} & {\bf Moderate} & {\bf Hard}\\ 
        \midrule
        LaserNet & 79.19 & 74.52  & 68.45 \\
        RCD (Ours) & 82.26 & 75.83 & 69.91 \\
        RCD-FT (Ours) & 85.37 & 82.61 & 77.80 \\
		\bottomrule\\[-1mm]
        \end{tabular}
	}
	\caption{Comparison to official LaserNet BEV results on KITTI Car testset. No test-time augmentation or additional training data is used for RCD. RCD method is trained from scratch on KITTI while RCD-FT is finetuned from WOD pretraining.}
	\label{tab:kitti_test}
\end{SCtable}

WOD captures multiple major cities in the U.S., under a variety of weather conditions and across different times of the day. The dataset provides a total number of 1000 sequences of 20s duration each, sampled at 10Hz, with train/validation split of 798/202 sequences.
The effective annotation radius for 6M vehicles across these sequences is 75 meters. For our experiments, we evaluate both Average Precision (AP) and AP weighted by heading (APH)~\cite{sun2019scalability} in 3D and BEV for vehicles on the WOD validation set and 3D detection metrics using the public evaluation server for the test set. 

\begin{table}[t]
    \centering
    \adjustbox{width=\textwidth}{
        \begin{tabular}{lcccccccccccccccc}
        \toprule
         & \multicolumn{4}{c}{3D AP (IoU=0.7)}     
         & \multicolumn{4}{c}{3D APH (IoU=0.7)}      
         & \multicolumn{4}{c}{BEV AP (IoU=0.7)}      
         & \multicolumn{4}{c}{BEV APH (IoU=0.7)}                  
         
         \\ \cmidrule(lr){2-5} \cmidrule(lr){6-9} \cmidrule(lr){10-13} \cmidrule(lr){14-17}
                    & All & {\small $r_{\leq 30}$}  & {\small $r_{30-50}$} & {\small $r_{\geq 50}$} & All & {\small $r_{\leq 30}$}  & {\small $r_{30-50}$} & {\small $r_{\geq 50}$} 
                    & All & {\small $r_{\leq 30}$}  & {\small $r_{30-50}$} & {\small $r_{\geq 50}$} & All & {\small $r_{\leq 30}$}  & {\small $r_{30-50}$} & {\small $r_{\geq 50}$}\\ 
        \midrule                    
        P.Pillars \cite{lang2019pointpillars} & 56.62 & 81.0 & 51.8 & 27.9 & - & - & - & -  & 75.57 & 92.1 & 74.1 & 55.5 & - & - & - & - \\
        
        DynVox \cite{zhou2019end} & 59.29 & 84.9 & 56.1 & 31.1 & - & - & - & -  & 77.18 & 93.0 & 76.1 & 57.7 & - & - & - & - \\ %
        
        MVF \cite{zhou2019end} & 62.93 & 86.3 & 60.0 & 36.0 & - & - & - & - & 80.40 & \second{93.6} & 79.2 & 63.1 & - & - & - & - \\
        
        PV-RCNN \cite{shi2019pv} & \best{70.30} & \best{91.9} & \best{69.2} & \second{42.2} & \best{69.69} & \best{91.3} & \best{68.5} & \second{41.3} & \second{82.96} & \best{97.4} & \best{83.0} & \second{65.0}  & \second{82.06} & \best{96.7} & \second{80.0} & \second{63.2} \\

        \midrule
        LaserNet*       & 52.11 & 70.9 & 52.9 & 29.6 & 50.05 & 68.7 & 51.4 & 28.6 & 71.19 & 83.9 & 71.4 & 54.5 & 67.66 & 80.7 & 68.4 & 51.4 \\  
        
        \textbf{RCD} (Ours)  & \second{69.59} &  \second{87.2} & \second{67.8} & \best{46.1} & \second{69.16} & \second{86.8} & \second{67.4} & \best{45.5} 
        & \best{83.35} & 93.5 & \second{82.3} & \best{67.9} & \best{82.64} & \second{93.0} & \best{81.6} & \best{66.6} \\  %
        
        \bottomrule\\[-1mm]
        \end{tabular}
    }
    \caption{Comparison of methods for vehicle detection on the Waymo Open Dataset (WOD) validation set for 3D detection with 7DOF boxes. The \best{best} and \second{second best} results are highlighted in bold and underlined respectively. (*) Our implementation of LaserNet \cite{meyer2019lasernet}. Columns with $r$ show breakdown of metrics by range (in meters).}
    \label{tab:eval_vehicle_3d_combi}
    \vspace{-3mm}
\end{table}

\subsection{Baseline Methods}
The full RCD model is compared to an equivalent two-stage detector baseline without SRG and RCD layers replaced with a fixed $7 \times 7$ dilated convolution (set to a dilation rate of 3 which is shown to have best RPN performance).
Furthermore, a selection of state of art methods from each mainstream category in 3D object detection algorithms are compared on WOD.

\textbf{LaserNet \cite{meyer2019lasernet}: } LaserNet is a 2D CNN-based singleshot 3D object detector operating on LiDAR range-images. It showed improvements on a large private dataset. With no publicly available implementation for this method, we use a variant of our RPN sub-network with normal 2D convolutions, ResNet blocks, adaptive NMS, and trained with multi-model box regression loss as described in~\cite{meyer2019lasernet}.

\textbf{Point Pillars (P.Pillars) \cite{meyer2019lasernet}: } Another single stage detector which utilizes PointNets \cite{qi2017pointnet} to encode a pointcloud scene representation organized in vertical columns in the BEV. Metrics from~\cite{zhou2019end}. 

\textbf{Multi-View Fusion (MVF) \cite{zhou2019end}: } This method fuses Cartesian view features and spherical view features. It shows significant improvements on long range detection because of the spherical view features. We share the same findings in our method as our method is perspective only.

\textbf{Point-Voxel (PV-RCNN) \cite{shi2019pv}: } a recently proposed method combining PointNets \cite{qi2017pointnet++} and a sparse convolution RPN backbone with a similar second stage RCNN refinement network. %

\subsection{Implementation details}

For the RCD layer we always use $N=64$ number of samples initialized as a $8 \times 8$ grid. With the number of channels kept at 64 for the inputs and outputs of the RCD block, the entire block consumes 23K FLOPs per pixel which is significantly lower than 262K FLOPs for an equivalent 2D convolutional layer, primarily due to PConv squeeze with only 3 filters. Our RPN network takes 74ms per frame compared to 301ms for our best-effort implementation of the RPN from PV-RCNN on a V100 GPU.
All validation experiments use the Adam optimizer \cite{kingma2015adam} for 350K iterations of batch-size 8 (or 17.5 epochs) with a learning rate starting from 6e-3 with a cosine decay end to end from scratch without any data augmentation. For submission to the WOD test-server our RCD model is trained for a total of 1 million iterations.

\subsection{Discussion of Results}
Among the baseline methods, only LaserNet~\cite{meyer2019lasernet} uses the range imagery directly, this makes LaserNet our direct comparison.
The results of this are shown in Table~\ref{tab:eval_vehicle_3d_combi}.
Even without RCNN our RPN achieves 57.2 AP, a significant improvement over the LaserNet range image detector.
This is mainly because of the way RCD and SRG handle scale variance and occlusion in the range image view. 
As our method processes the range image in the perspective view it has fundamentally different characteristics compared to voxel or BEV projection based methods.
Compared with the voxel based methods reported on WOD, our method shows greatest improvements in long range. This is due to the issue around voxel sparsity for distant objects corroborating with the findings of \cite{zhou2019end}. Additionally, RCD is better able to utilize contextual information in the range image to distinguish distant objects with few points.
As a result of this, our method exhibits complementary performance to the state-of-the-art PV-RCNN method which has strongest performance in close range where point density is highest for voxel based detection.

\begin{SCtable}
	\adjustbox{width=.45\textwidth}{
		\begin{tabular}{lcccc}
			\toprule
			& \multicolumn{2}{c}{Level 1} & \multicolumn{2}{c}{Level 2}               \\ 
			\cmidrule(lr){2-3} \cmidrule(lr){4-5}
			& AP & APH  & AP & APH   \\ 
			\midrule                    
			Second \cite{yan2018second} & 50.11 & 49.63 & 42.88 & 42.48  \\ %
			P.Pillars \cite{lang2019pointpillars} & 54.94&  54.47 & 48.61 & 48.18  \\ %
			StarNet \cite{ngiam2019starnet}& 61.68 &  61.23 & 55.17  & 54.76  \\ 
			SA-SSD \cite{he2020sassd}& \second{70.24} &  \second{69.54} & \second{61.79}  & \second{61.17}  \\ 
			\midrule
			RCD 1M (Ours)  & \best{71.97}  &  \best{71.59} & \best{65.06}  &  \best{64.70} \\
			
			\bottomrule\\[-1mm]
		\end{tabular}
	}
	\caption{Comparison of single frame methods for 3D vehicle detection on the Waymo Open Dataset (WOD) test set. Values are provided by the test server and divided into two levels of difficulty where Level 1 has at least five points per ground truth object and Level 2 can have as few as a single point.}
	\label{tab:eval_vehicle_3d_testset}
\end{SCtable}

\textbf{WOD Leaderboard}:\quad Recently, an unlabelled test set was released to the public by Waymo for benchmarking 3D object detectors via an online evaluation service. Table \ref{tab:eval_vehicle_3d_testset} shows the performance of our best model trained for 1 million iterations compared to other published single LiDAR frame vehicle detectors on the leaderboard\footnote{\href{https://waymo.com/open/challenges/3d-detection/}{waymo.com/open/challenges/3d-detection/}, results gathered on July 27$^{th}$, 2020}. The public leaderboard divides the detection results into two difficulty levels based on the number of points within the annotated boxes. Level 1 has at least five points per ground truth object and Level 2 boxes may only have a single point. Our range based RCD model significantly outperforms other methods for both difficulty levels on this public benchmark.

\textbf{Ablation Study:}
The merits of the proposed RCD block is compared with standard dilated convolution.
Furthermore, the proposed RCD layer relates to the exhausted dilation combination of differently sized receptive kernels in ASPP~\citep{chen2017deeplab}. The key difference is that ASPP merges multiple fixed dilated convolutions together while RCD reuses the same set of filter weights over a continuum of dilations. We found that substituting our RCD for an ASPP block often resulted in overfitting or unstable training with only a single variant producing comparable results shown as the third column of results in Table \ref{tab:ablation_study}. As ASPP uniformly applies multiple rates of dilation simultaneously its large receptive field is susceptible to distracting object in the periphery.
Using the RCD with SRG at the start and throughout the backbone network achieves the best single stage performance while ASPP exhibited the worst performance. The last three columns show the effect of the two-stage framework. Interestingly, the 3D RCNN second stage is able to substantially recover from the under-performing performing ASPP based RPN. The RCNN also capitalizes on the high-quality proposals from the RPN network endowed with multiple RCD blocks.

\begin{table}[t]
    \centering
    {
    \adjustbox{width=.85\textwidth}{
    \begin{tabular}{lccccccccc}
        \toprule
        Fixed dilation (rate=3) & & \checkmark &  &  & & & &  \checkmark & \\
        ASPP && &  \checkmark&&&&&&  \checkmark  \\
        RCD at start &&&& \checkmark & \checkmark & \checkmark & \checkmark & & \\
        With SRG &&&&&  \checkmark & \checkmark & \checkmark & & \\
        RCD at multiple scales &&&&&& \checkmark & \checkmark & & \\
        3D RCNN &&&&&&& \checkmark & \checkmark  & \checkmark \\
        \midrule
         3D AP   & 52.3  & 53.8  & 45.8 &  54.9     & 55.0   &  57.2   & \best{69.6} & 63.6  & \second{64.4} \\
         3D APH  & 50.3  & 51.8  & 45.2 &  52.5     & 52.9   &  56.7   & \best{69.2} & 63.1  & \second{63.9} \\
        \bottomrule\\[-1mm]
    \end{tabular}
    }
    }
    \caption{Ablation study showing the progressive improvements with proposed contributions. Measured 3D AP and APH are from the WOD validation sequences. RCD at multiple scales uses 4 blocks with one at the start of the RPN and others at downsampled resolutions (see supplementary).}
    \label{tab:ablation_study}
    \vspace{-3mm}
\end{table}

\section{Conclusions}

We introduce RCD, a method for dynamically adjusting the dilation rate for use with LiDAR range images for scale invariant 3D object detection.
An improved system relying on our new RCD block representation, and based on the two stage RCNN method \cite{shi2019pointrcnn}, is the top performing range image based detection method, over all ranges, on the Waymo Open Dataset \cite{sun2019scalability} benchmark for BEV and achieves competitive results in other tests.
Specifically, our approach sets a new state-of-the-art for detecting vehicles at long distances as it benefits from the dense nature of the range image.
Two directions of future research include, applying the proposed method with other common robotic depth sensors, like structured-light or time-of-flight cameras which also capture a 2D range image, and designing a hybrid PV-RCNN and RCD approach to efficiently obtain the best performance at all ranges.

\if\ShowAppendix1
\clearpage
\appendix

\section{Further Implementation Details}

\subsection{Input Range Image}
The input range image format to the entire network follows what is provided by the Waymo Open
Dataset (WOD) \cite{sun2019scalability} with rows corresponding to LiDAR beam inclinations and columns corresponding to laser shot azimuth. The range image pixels encoded channels include: range, intensity, elongation \cite{sun2019scalability, shand2020el}, inclination, azimuth, x, y, z. 
Most of these channels are defined clearly in the WOD \cite{sun2019scalability}. The channels x, y, z are the Cartesian coordinates of each point in the vehicle frame.

\subsection{RCD-RPN Backbone Network}

As briefly described in the main paper the backbone of the RCD-RPN is a 2D convolutional network inspired by the deep layer aggregation network \cite{yu2018deep} used in LaserNet\cite{meyer2019lasernet}. The architecture is composed of ResNet convolutional bottleneck units~\cite{he2016deep} with multiple steps of horizontal downsampling as shown in the upper blue section of Figure~\ref{fig:both_stages} (see caption for the number of ResNet units per downsampled scale). The aggregator modules take in two feature maps at different resolutions, upsampling the lower resolution via transposed convolutions. 

Beyond substituting an initial convolution applied to the LiDAR input,
the RCD block can be applied anywhere within the convolutional network.
For example, replacing a ResNet block with an RCD block is considered at Res1, Res2, and Res3 of our multi-resolution backbone. The range image used to condition the amount of dilation at each pixel is coarsely downsampled using max-pooling, with the intention to favour distant objects with fewer pixels.

\begin{figure*}[h]
	\centering
	\includegraphics[width=.95\textwidth]{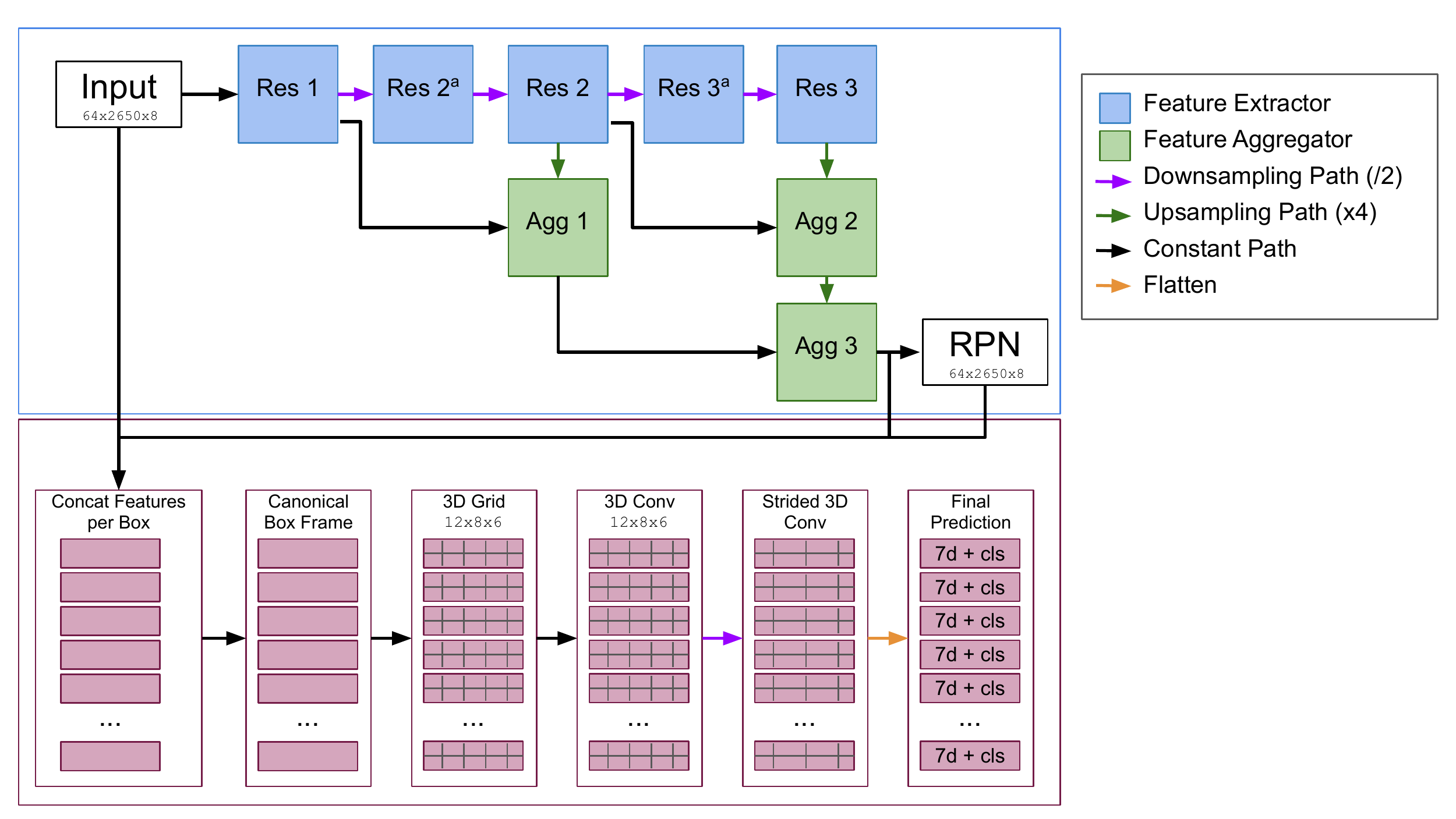}
	\caption{Illustrative detail of our detection pipeline. \textbf{RPN:} The upper blue section shows our 2D CNN backbone where the Resnet blocks and aggregation blocks contain either RCD layers or convolutational layers for the baseline. Number of bottlenecks units in each block: (Res1: 5), (Res2,Res2a: 7), (Res3,Res3a: 9), (Agg1,Agg2,Agg3: 4). Strides in each block: (Res1: 2), (Res2, Res2a: 2), (Res3, Res3a: 2). The bottleneck is the one described in~\cite{he2016deep} with bottleneck depth set to the bottleneck input channel size divided by 4. The output channel size from each layer are 64, 128, 256 corresponding to the block numbers. The final RPN head predicts box parameters and score. \textbf{RCNN:} The second stage operates independently per surviving box. The Cartesian point coordinates are combined with box parameters to transform features range-image into their box canonical frame.  
	For each box, point features are voxelized and followed by 3D convolutions and final predictions as shown.} 
	\label{fig:both_stages}
\end{figure*}

\section{Learnable Sampling Pattern Analysis}

To further assess the effect of using a learnable, continuous dilated kernel sampling, Figure \ref{fig:pattern_movement}  shows the relative locations of the learnt spatial samples with their initial positions marked. Compared to the uniform sampling at initialization, inner sample points tend to increase the local concentration around the center with the outer points showing negligible or slight spread away from the central location. This indicates that the RCD model attempts to over-sample in the central region with sparser sampling in the extremities, similar to the Fovea in the human eye. %
Figure \ref{fig:nominal_width_plot} plots the evolution of the learnable nominal width parameter $\lambda$ from Equation %
2 of the main paper.

\noindent\textbf{Effect of Nominal Width}:\quad For assessing the behaviour of the nominal width in all experiments we set it as a learnable parameter initialized to the value of 1m. Over the course of training, this parameter generally increases and then settles between 2 and 3m which is comparable to the average dimensions of a vehicle. 

\vspace{2mm}
\noindent\textbf{Effect of SRG}:\quad We found that the addition of a range gate improved training stability leading to faster convergence with a slightly improved performance.
Table~\ref{tab:compare_to_standard_supp} shows the benefits provided to the first stage detector compared to adding a standard square kernel with different dilation rates. For a fair comparison we substitute the RCD block with a standard 2D convolutional layer with $7 \times 7$ kernel and 64 channels matching the RCD output.

\begin{figure}
    \centering
    \begin{minipage}[b]{0.45\textwidth}
        \includegraphics[width=\textwidth]{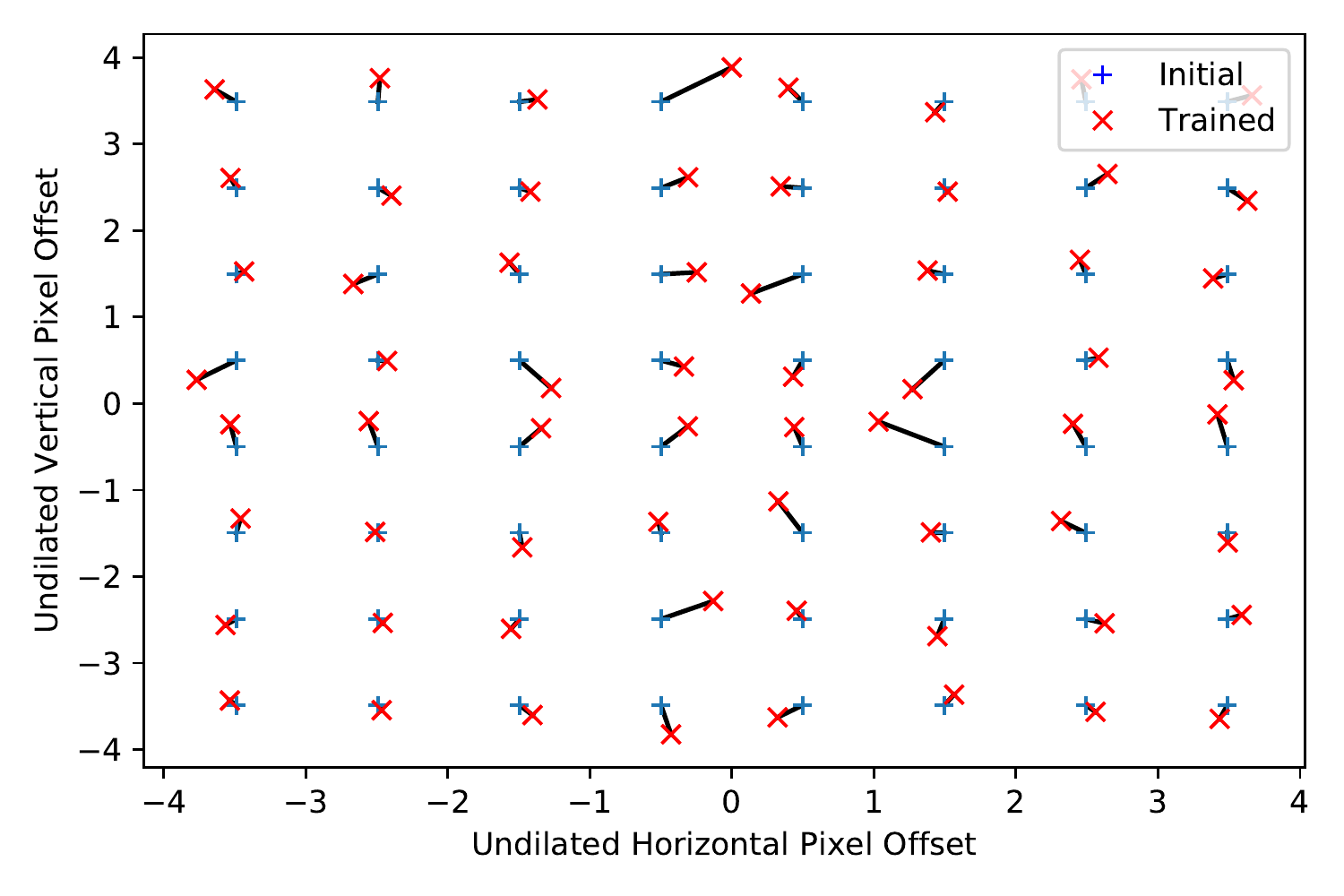}
	    \caption{Movement of a $8 \times 8$ kernel sampling pattern over the course of training. With blue `\textcolor{blue}{$+$}' showing the uniform grid used to initialize offsets, and red `\textcolor{red}{$\times $}' showing the pattern after training.}
        \label{fig:pattern_movement}
    \end{minipage}%
    \qquad
    \begin{minipage}[b]{0.45\textwidth}
        \includegraphics[width=\textwidth]{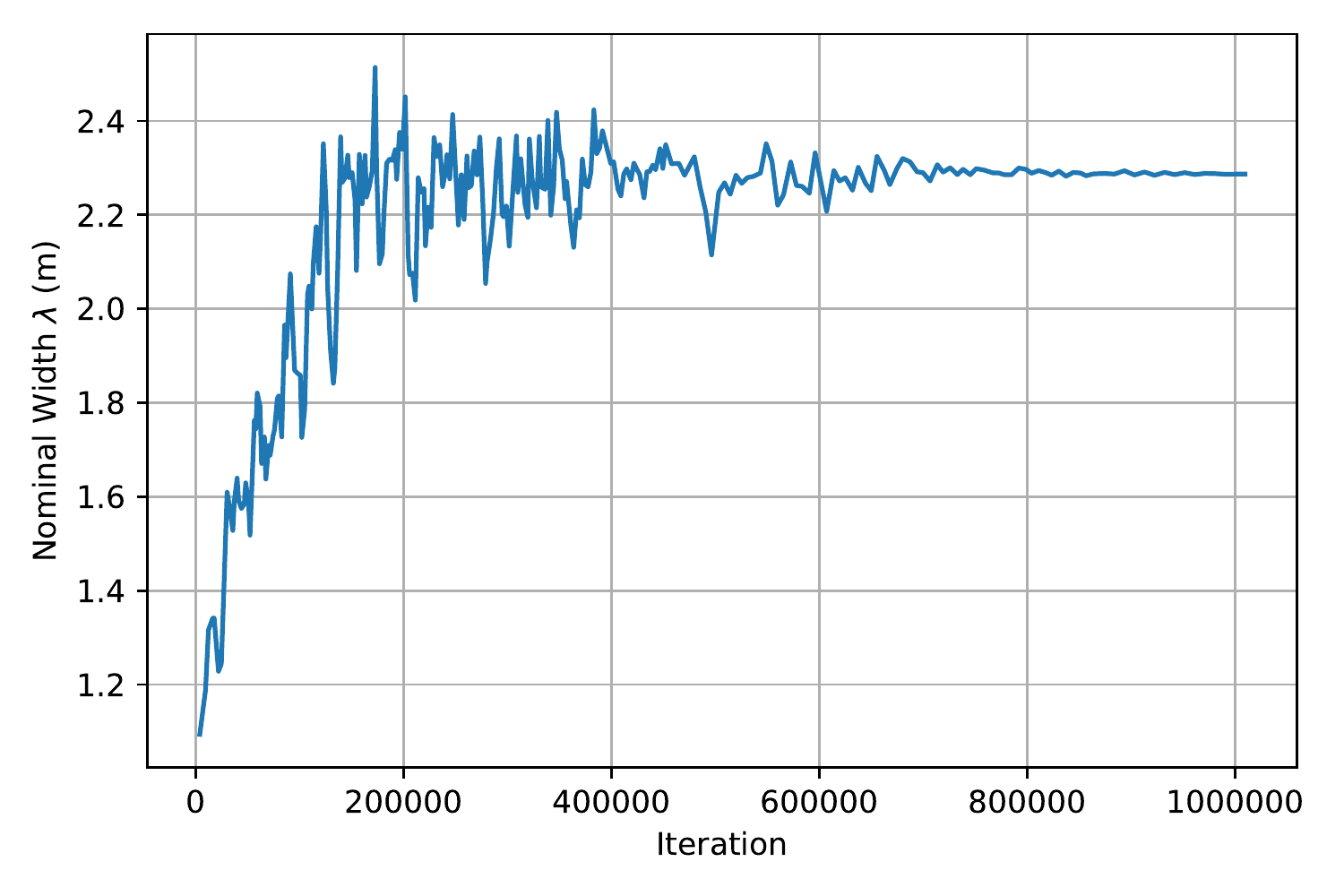}
    	\caption{Nominal width free parameter $\lambda$ over the course of training for the input RCD block. $\lambda$ is used to model the nominal width (in meters) of an object at any range. }
        \label{fig:nominal_width_plot}
    \end{minipage}%
\end{figure}

\begin{table}
    \centering 
    \begin{tabular}{lcccc}
        \toprule
        Method & Dilation & AP & APH & R@50P\\
        \midrule
        \multirow{4}{*}{Standard Convolution} & Fixed rate=1 &  52.25 & 50.28 & 59.22 \\
        & Fixed rate=3 & 53.75 & 51.76 & 60.05 \\
        & Fixed rate=5 & 53.61 & 51.3 & 59.93 \\
        & Fixed rate=7 & 51.58 & 48.06 & 60.23 \\
        \midrule
        ASPP*~\cite{chen2017deeplab} & (12, 24, 36, G) & 45.8 & 45.2 & - \\
        \midrule
        Ours & Range conditioned & 54.87 & 52.47 & \best{61.03} \\
        Ours with SRG & Range conditioned & \best{55.01} & \best{52.89} & 60.83 \\
        \bottomrule\\[-1mm]
    \end{tabular}
    \caption{Performance comparison for first stage RPN with a single RCD or standard convolution applied to the input tensor with fixed dilation rates. Average precision (AP), average precision weighted by heading (APH) and Recall at 50\% precision (R@50P) is measured on WOD validation set after 350k iterations of training. (*) ASPP - with 16 filter channels per dilation layer - peaked near 240K iterations with values shown in table.
    RCD improves over using only standard convolutional layers, and adding soft range gating (SRG), improves AP and APH even further. 
    }
    \vspace{-2mm}
    \label{tab:compare_to_standard_supp}
\end{table}

\begin{table*}[t]
	\centering
	
	\adjustbox{width=.8\textwidth}{
        \begin{tabular}{ccccccc}
        \toprule
        \# Channels & Dilations & RPN-AP & RPN-APH &  RCNN-AP & RCNN-APH & Peak@iteration \\ 
        \midrule
        64 &	(2, 4, 8, G) &	28.6 &	28.0 &	59.2 &	58.7 &	83K \\
        64 &	(6, 12, 18, G) &	27.1 &	26.4 &	58.4 &	58.0 &	62K\\
        64 &	(12, 24, 36, G) &	30.4 &	29.6 &	58.7 &	58.2 &	60K\\
        16 &	(12, 24, 36, G) &	45.8 &	45.2 &	64.4 &	63.9 &	240K\\
        8 &	(12, 24, 36, G) &	27.5 &	27.0 &	59.4 &	59.0 &	76K\\
		\bottomrule\\[-1mm]
        \end{tabular}
	}
	\caption{Parameter search for the ASPP~\cite{chen2017deeplab} baseline where all methods were trained for 350K iterations. Peak@iteration shows the checkpoint which the highest second stage RCNN performance on WOD validation. The low and sporadic RPN performance and early peak iteration suggests that ASPP easily overfits and generally results in unstable training.}
	\label{tab:aspp_search}
    
\end{table*}

\section{Comparison to Fixed Dilation Rates}
Table \ref{tab:compare_to_standard_supp}, expands the experiments in the main paper with a more exhaustive set of dilation rates. Here we explore the effect of increasing the dilation rate of the kernel for a standard convolutional operation \cite{chen2017deeplab,yu2015multi} in the first region proposal network (RPN). 
We also compare to the atrous spatial pyramid pooling (ASPP)~\cite{chen2017deeplab} framework by replacing our RCD layer with the ASPP module. Here we used fixed strides of 12, 24, 36 and global average features denoted by `G'\footnote{Following implementation of \cite{chen2017rethinking} from: github.com/rishizek/tensorflow-deeplab-v3}.

With dilation rates of 3 and 5, a small improvement over the standard convolution (with default dilation rate of 1) is observed. However, further increasing the dilation results in a sharp drop in performance as seen with the dilation rate of 7. This drop in performance could be due to: insufficient sampling of small distant objects; or an increase in the amount of padding needed given the height of the range image is limited to 64 pixels. ASPP with various dilations significantly under-performed compared to the standard dilated convolutions. Having fixed and overly large dilation rates increases the susceptibility of ASPP to be distracted by sporadic activations caused by high frequency detail. We tried to vary the receptive field and the number of channels (parameters) to prevent overfitting but found the training to be generally highly unstable. See Table~\ref{tab:aspp_search}. The RCD models, with their ability to appropriately adjust their rate of dilation for both near and far, leads to the best overall performance.

\section{Qualitative Results}

Figures \ref{fig:scene_viz} and \ref{fig:bev_rpn_rcnn_gt} provide a qualitative view of our detections on WOD validation sequences. In top left of the third row of Figure \ref{fig:bev_rpn_rcnn_gt} the RPN conservatively predicts a proposal in the occlusion shadow of a vehicle which is later correctly removed by the second stage RCNN refinement. A failure case is shown in the first row, where the RCNN mistakenly classifies a proposal as background. %

\begin{figure}[]
	\centering
	\includegraphics[width=0.16\textwidth]{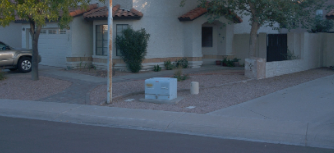}
	\includegraphics[width=0.16\textwidth]{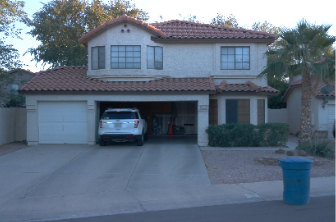}
	\includegraphics[width=0.16\textwidth]{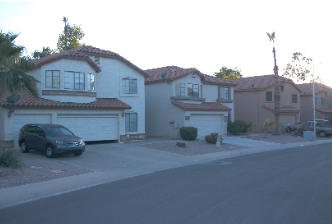}
	\includegraphics[width=0.16\textwidth]{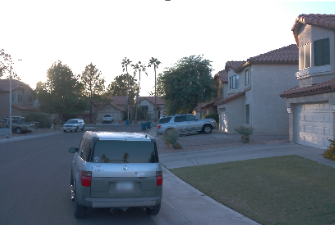}
	\includegraphics[width=0.16\textwidth]{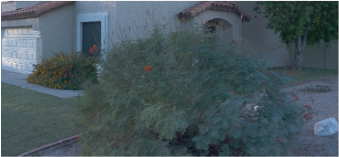}
	\\
	\includegraphics[width=0.8\textwidth]{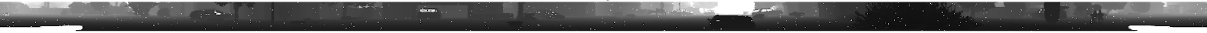}
	\\
	\includegraphics[width=0.8\textwidth]{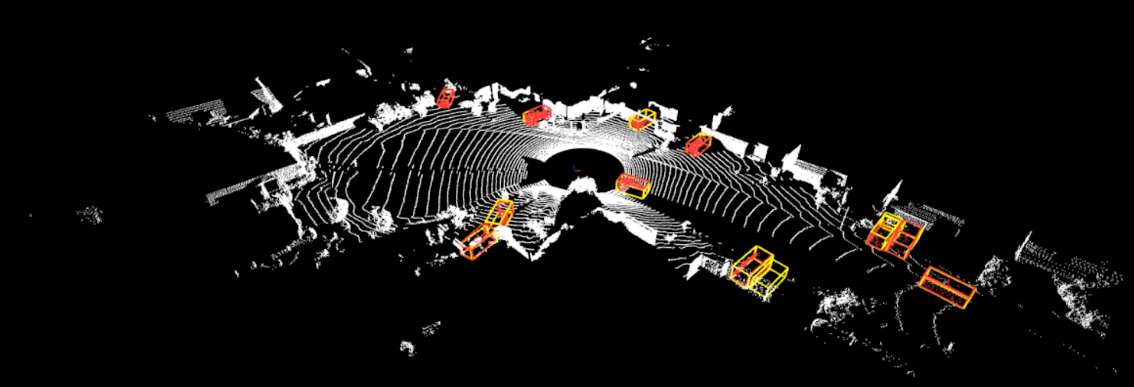}
	
	\caption{Example visualization of 3D alignment. Top row shows colour images of Left, Front-Left, Front, Front-Right and Right view respectively. Middle row shows the corresponding range image for the scene. Bottom row shows the corresponding 3D pointcloud with our detections in red against the ground truth boxes in yellow. \emph{Note the RGB images are only for illustration purposes.}}
	\label{fig:scene_viz}
\end{figure}

\section{Comparison to LaserNet on KITTI}

\begin{figure}[ht]
    \centering
    \includegraphics[width=.8\textwidth]{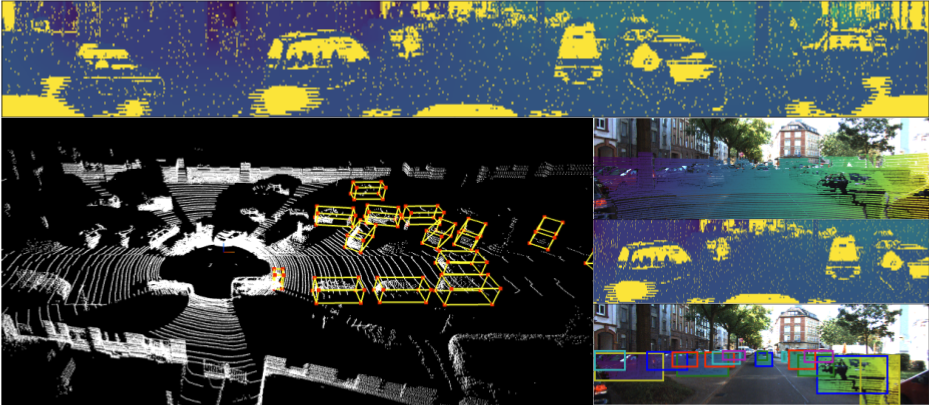}
    \caption{Illustration of extracted range image (\textbf{top}) from KITTI point cloud (\textbf{left}), with projection of the point cloud onto the RGB image (\textbf{middle right}), a detail of the extracted range image, and (\textbf{bottom right}) an illustration of the visible points per annotated bounding box.}
    \label{fig:range_vis}
\end{figure}

While the KITTI dataset is a common benchmark for 3D detection, it has been shown that its small size can lead to overfitting and poor generalization for range image based methods~\cite{meyer2019lasernet}. While we also observe similar issues, the KITTI dataset is used as an indication of how our method compares to the official results of LaserNet in the small dataset regime.

\subsection{Range Images from Point Clouds}
Despite the range-image being a native representation for LiDAR, KITTI distributes data from the Velodyne LiDAR sensor as a Cartesian pointcloud with an additional intensity channel.
The straightforward conversion of Cartesian to spherical points results in poor quality range-images due to the 64 lasers of the Velodyne not sharing a common origin. To overcome this issue, dense and compact range images are reconstructed as follows:
\begin{enumerate}
    \item For each of the lasers in the Velodyne sensor its inclination and height are determined using Hough voting, in a discretized height and inclination space. 
    \item For each point in a point cloud its corresponding laser-id $l$ and azimuth angle $a$ is estimated by minimizing the reconstruction / quantization error.
    \item The range image is then constructed as a $L \times A$ image, where $L$ denotes the number of lasers (64) and $A$ is the number of azimuth steps (set to 2048). The pixel value at location $(l,a)$ reflects the observed range, \ie $r = \sqrt{x^2 + y^2 + z^2}$, for point $p = (x, y, z)$). The addition of the laser intensity is similarly added as another channel. When multiple observations map to the same pixel location $(l, a)$ in the range image, the closest point is kept.
\end{enumerate}    
 Figure~\ref{fig:range_vis} illustrates the compact reconstructed range-image from KITTI Velodyne data.

\subsection{KITTI Results}
\label{sec:kitti_results}
We report performance on KITTI test set after a fixed number of 100K iterations, using
flipping and horizontal pixel shifting as data augmentation. Since the KITTI data is only labelled in the $90^\circ$ around the forward driving direction, only that part of the range image is fed into the network. Table~\ref{tab:kitti_test} shows the performance of our method against the primary representative for range-image detectors LaserNet~\cite{meyer2019lasernet} on this small dataset. This table is extended with a RCD model pretrained on WOD and finetuned on KITTI for 100K iterations, denoted as RCD-FT. While these numbers are not intended to be directly compared to other published works, they serve to high-light the importance of large and diverse datasets for training range-image detectors.

\begin{figure}[t]
	\centering
	\subfloat[Proposals]{
    \begin{minipage}[t]{0.32\textwidth}
	\centering
    	\includegraphics[width=\textwidth]{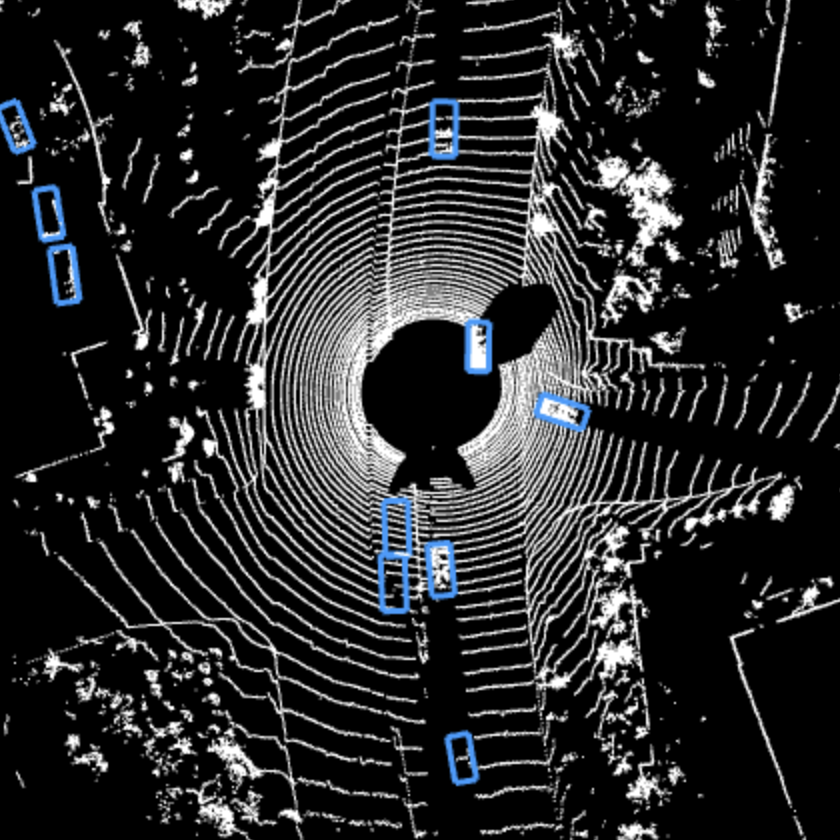}
    	~
    	\includegraphics[width=\textwidth]{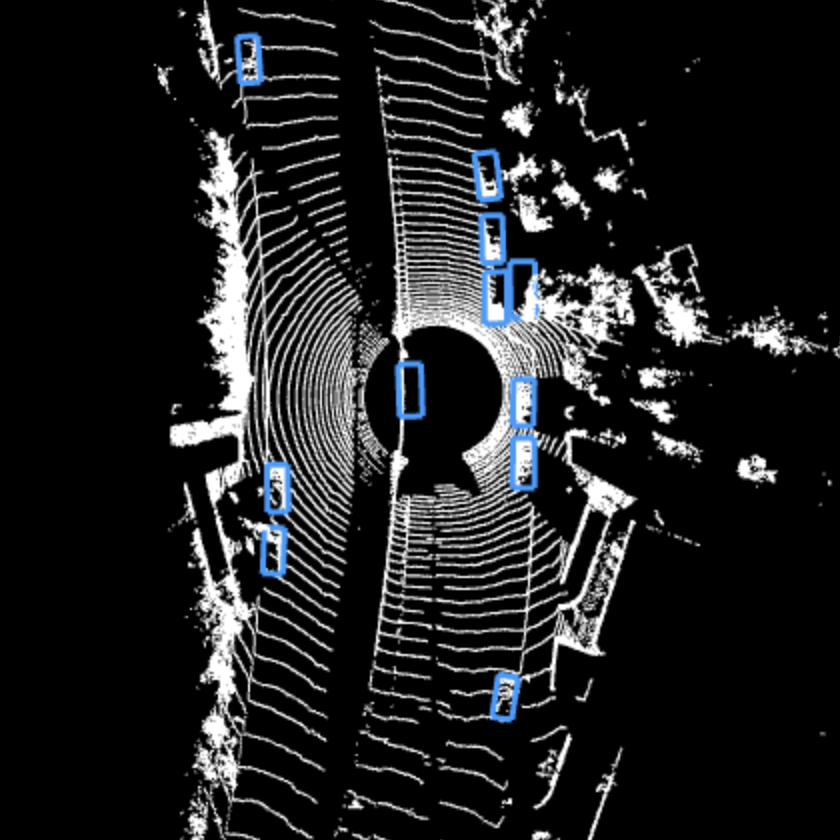}
    	~
    	\includegraphics[width=\textwidth]{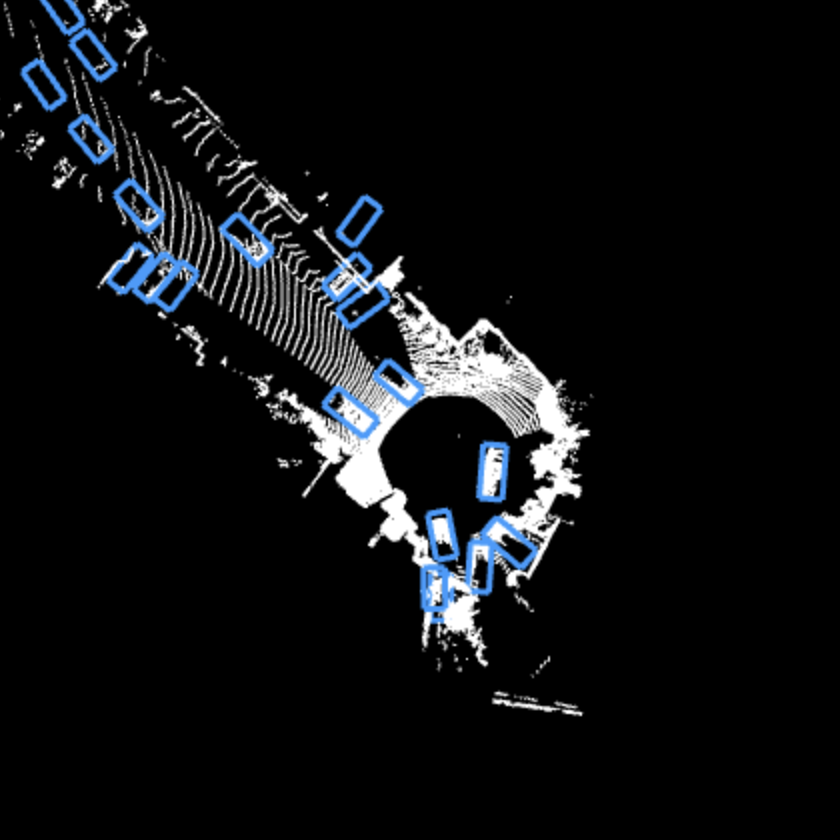}
    	~
    	\includegraphics[width=\textwidth]{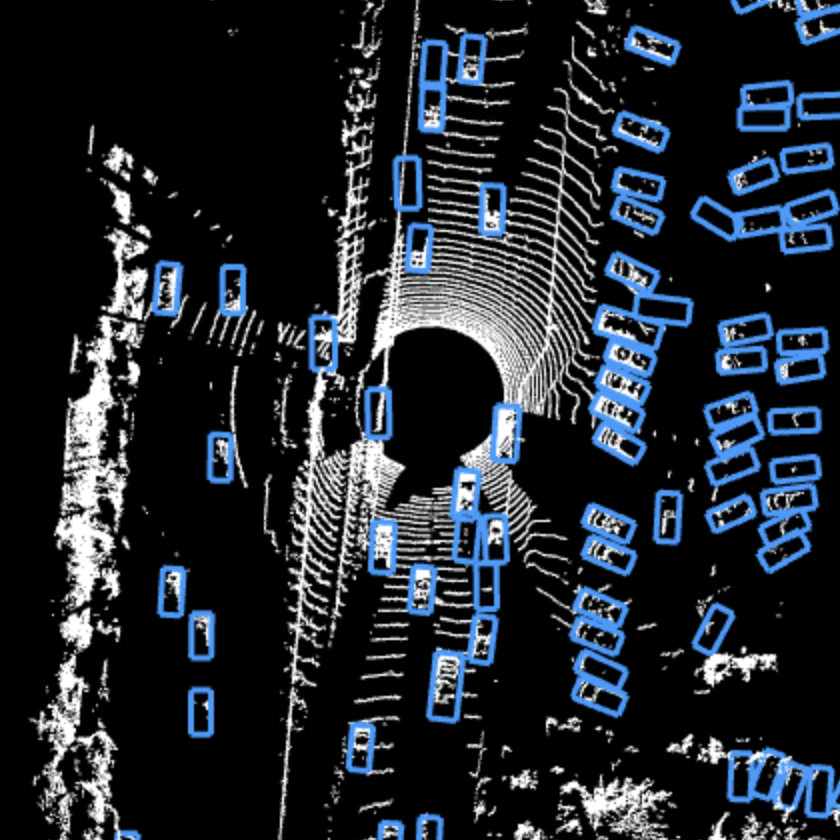}
    \end{minipage}}\hfill
    \subfloat[Detections]{
    \begin{minipage}[t]{0.32\textwidth}
	\centering
    	\includegraphics[width=\textwidth]{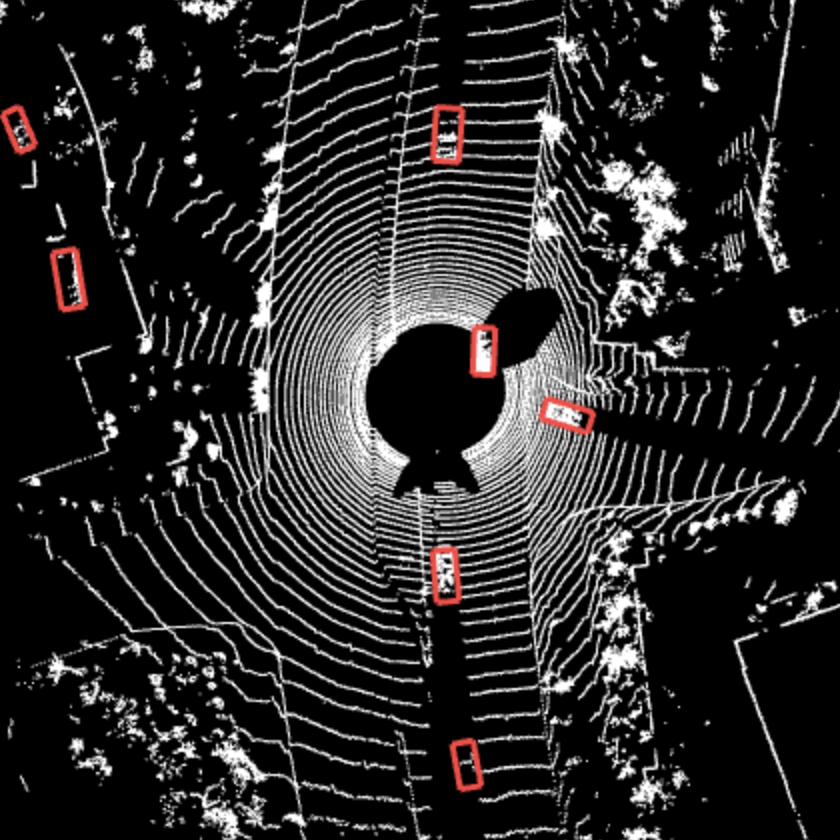}
    	~
    	\includegraphics[width=\textwidth]{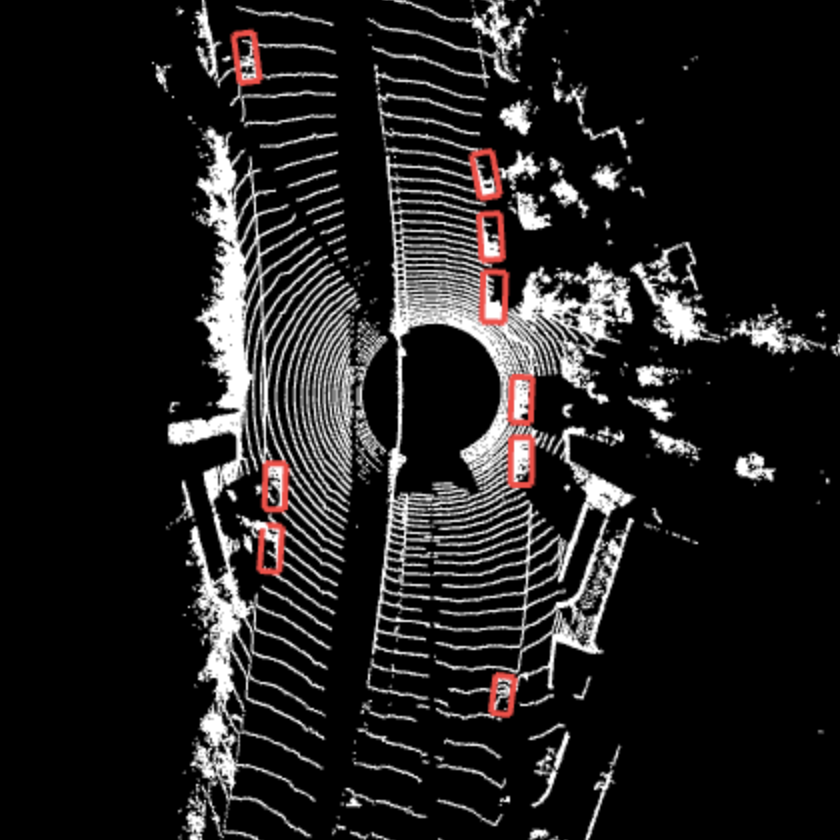}
    	~
    	\includegraphics[width=\textwidth]{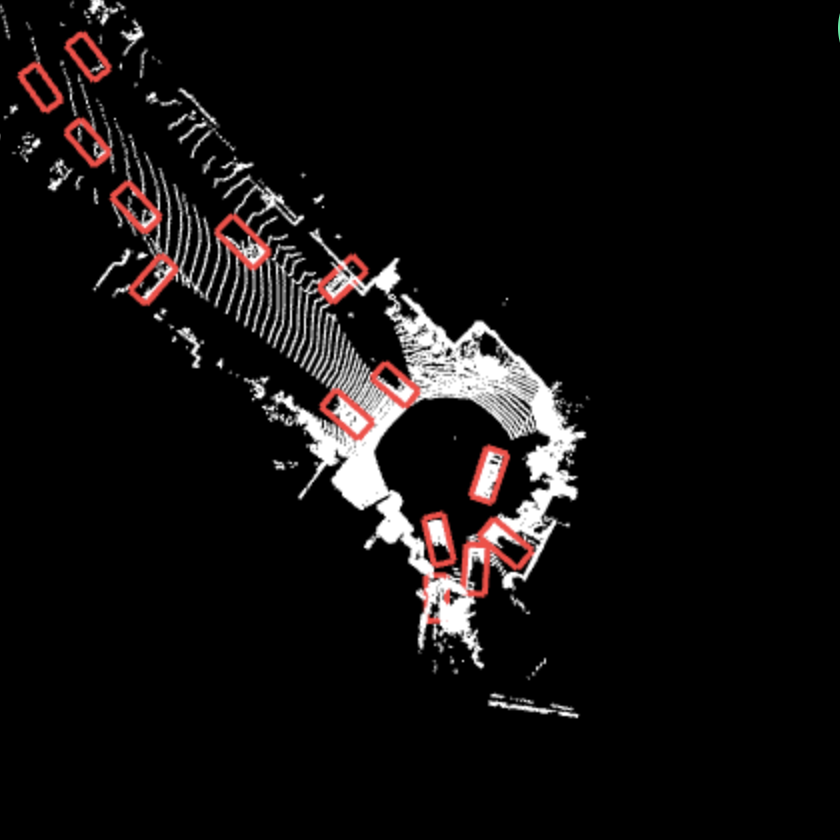}
    	~
    	\includegraphics[width=\textwidth]{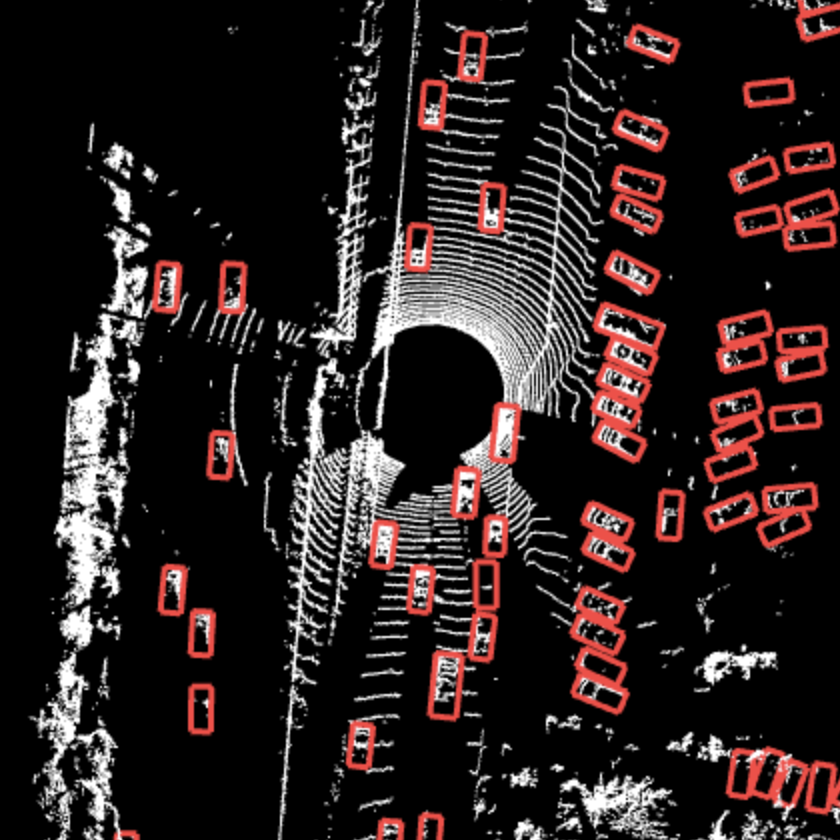}
    \end{minipage}}\hfill
    \subfloat[Ground Truth]{
    \begin{minipage}[t]{0.32\textwidth}
	\centering
    	\includegraphics[width=\textwidth]{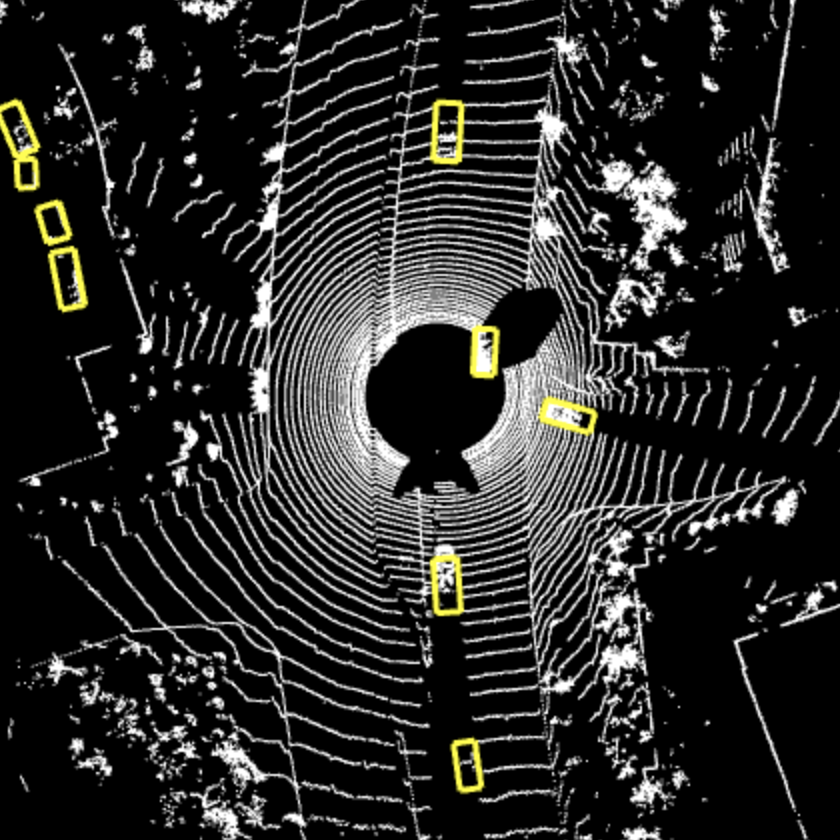}
    	~
    	\includegraphics[width=\textwidth]{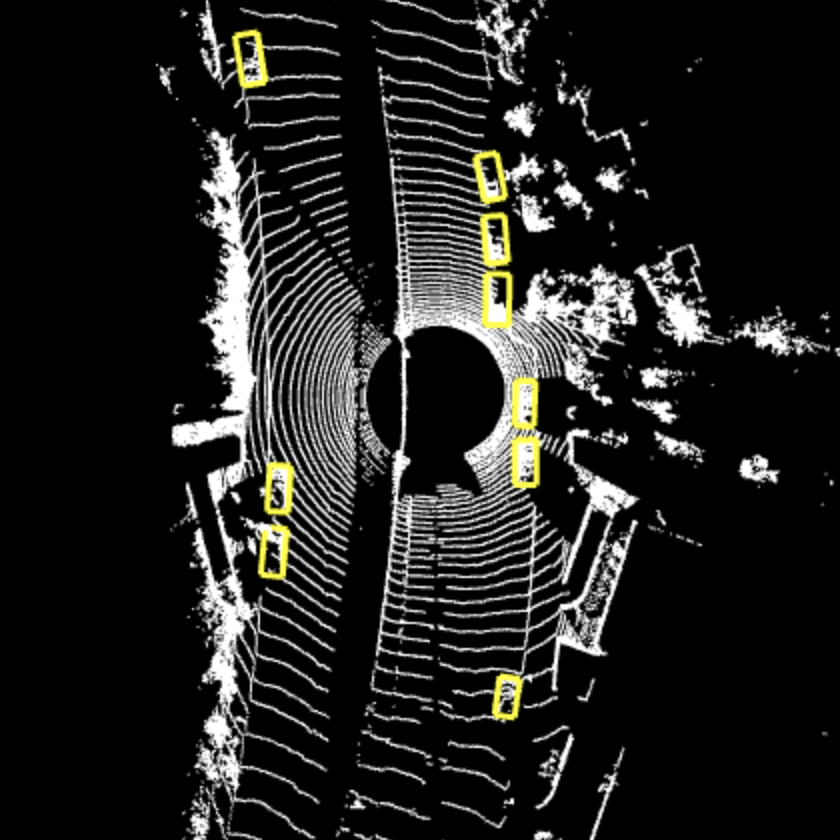}
    	~
    	\includegraphics[width=\textwidth]{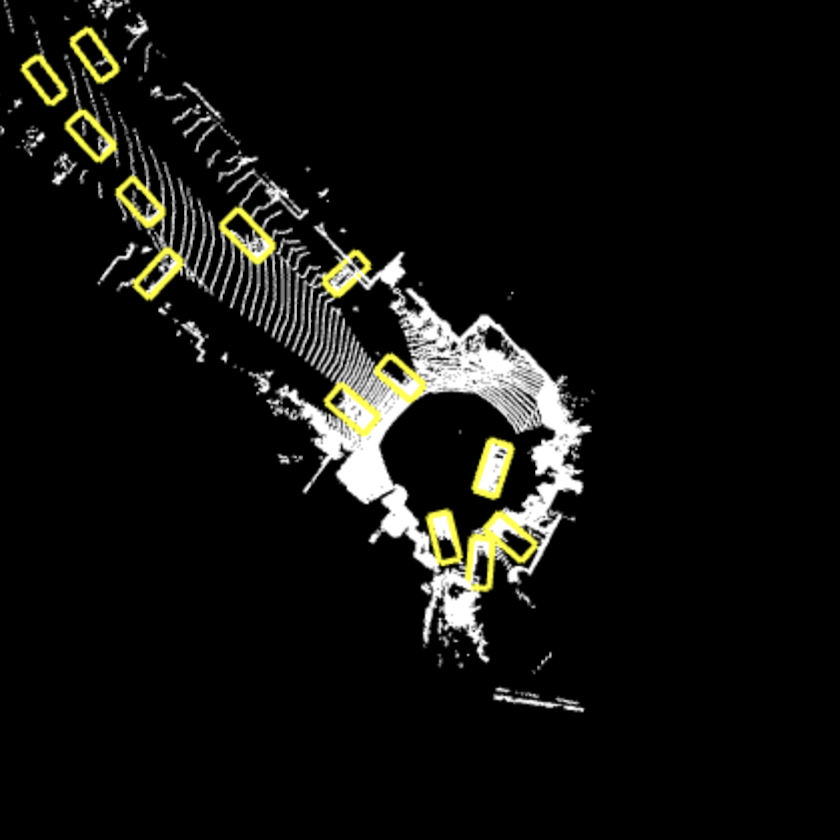}
    	~
    	\includegraphics[width=\textwidth]{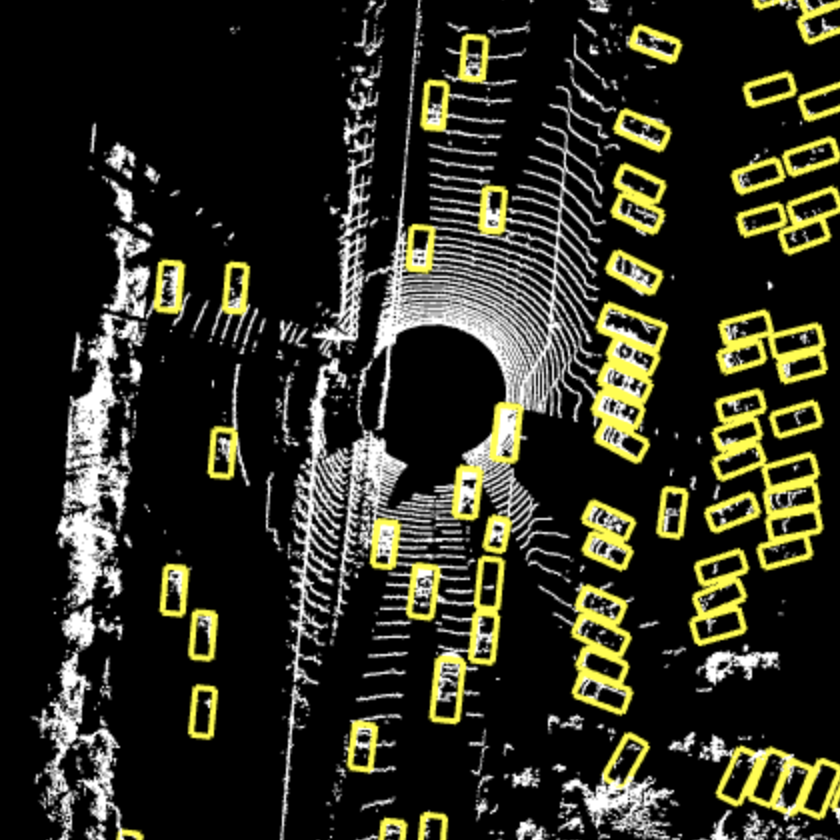}
    \end{minipage}}
	
	\caption{Birds-eye-view visualization of four different scenes from the WOD validation sequences. Each row shows a different scene and each column from left to right shows: output of the RPN in \textcolor{blue}{blue}, output of RCNN in \textcolor{red}{red}, and ground truth boxes in \textcolor{orange}{yellow} %
	respectively. All predictions are filtered with minimum confidence of $0.5$. Best viewed digitally with zoom.}
	\label{fig:bev_rpn_rcnn_gt}
\end{figure}

\fi


\begin{thebibliography}{51}
	\providecommand{\natexlab}[1]{#1}
	\providecommand{\url}[1]{\texttt{#1}}
	\expandafter\ifx\csname urlstyle\endcsname\relax
	\providecommand{\doi}[1]{doi: #1}\else
	\providecommand{\doi}{doi: \begingroup \urlstyle{rm}\Url}\fi
	
	\bibitem[Engelcke et~al.(2017)Engelcke, Rao, Wang, Tong, and
	Posner]{engelcke2017vote3deep}
	M.~Engelcke, D.~Rao, D.~Z. Wang, C.~H. Tong, and I.~Posner.
	\newblock Vote3deep: Fast object detection in 3d point clouds using efficient
	convolutional neural networks.
	\newblock In \emph{ICRA}. IEEE, 2017.
	
	\bibitem[Wang and Posner(2015)]{wang2015voting}
	D.~Z. Wang and I.~Posner.
	\newblock Voting for voting in online point cloud object detection.
	\newblock In \emph{Robotics: Science and Systems}, 2015.
	
	\bibitem[Zhou and Tuzel(2018)]{zhou2018voxelnet}
	Y.~Zhou and O.~Tuzel.
	\newblock Voxelnet: End-to-end learning for point cloud based 3d object
	detection.
	\newblock In \emph{CVPR}, pages 4490--4499, 2018.
	
	\bibitem[Chen et~al.(2017)Chen, Ma, Wan, Li, and Xia]{Chen_2017_CVPR}
	X.~Chen, H.~Ma, J.~Wan, B.~Li, and T.~Xia.
	\newblock Multi-view 3d object detection network for autonomous driving.
	\newblock In \emph{CVPR}, 2017.
	
	\bibitem[Lang et~al.(2019)Lang, Vora, Caesar, Zhou, Yang, and
	Beijbom]{lang2019pointpillars}
	A.~H. Lang, S.~Vora, H.~Caesar, L.~Zhou, J.~Yang, and O.~Beijbom.
	\newblock Pointpillars: Fast encoders for object detection from point clouds.
	\newblock In \emph{CVPR}, pages 12697--12705, 2019.
	
	\bibitem[Qi et~al.(2017{\natexlab{a}})Qi, Su, Mo, and Guibas]{qi2017pointnet}
	C.~R. Qi, H.~Su, K.~Mo, and L.~J. Guibas.
	\newblock Pointnet: Deep learning on point sets for 3d classification and
	segmentation.
	\newblock In \emph{CVPR}, pages 652--660, 2017{\natexlab{a}}.
	
	\bibitem[Qi et~al.(2017{\natexlab{b}})Qi, Yi, Su, and Guibas]{qi2017pointnet++}
	C.~R. Qi, L.~Yi, H.~Su, and L.~J. Guibas.
	\newblock Pointnet++: Deep hierarchical feature learning on point sets in a
	metric space.
	\newblock In \emph{NeurIPS}, 2017{\natexlab{b}}.
	
	\bibitem[Engelmann et~al.(2020)Engelmann, Kontogianni, and
	Leibe]{Engelmann20ICRA}
	F.~Engelmann, T.~Kontogianni, and B.~Leibe.
	\newblock {Dilated Point Convolutions: On the Receptive Field Size of Point
		Convolutions on 3D Point Clouds}.
	\newblock In \emph{{ICRA}}, 2020.
	
	\bibitem[Bewley and Upcroft(2013)]{bewley2013advantages}
	A.~Bewley and B.~Upcroft.
	\newblock Advantages of exploiting projection structure for segmenting dense 3d
	point clouds.
	\newblock In \emph{Australian Conference on Robotics and Automation}, 2013.
	
	\bibitem[Wu et~al.(2018)Wu, Wan, Yue, and Keutzer]{wu2018squeezeseg}
	B.~Wu, A.~Wan, X.~Yue, and K.~Keutzer.
	\newblock Squeezeseg: Convolutional neural nets with recurrent crf for
	real-time road-object segmentation from 3d lidar point cloud.
	\newblock In \emph{ICRA}, 2018.
	
	\bibitem[Meyer et~al.(2019)Meyer, Laddha, Kee, Vallespi-Gonzalez, and
	Wellington]{meyer2019lasernet}
	G.~P. Meyer, A.~Laddha, E.~Kee, C.~Vallespi-Gonzalez, and C.~K. Wellington.
	\newblock Lasernet: An efficient probabilistic 3d object detector for
	autonomous driving.
	\newblock In \emph{CVPR}, 2019.
	
	\bibitem[Sun et~al.(2020)Sun, Kretzschmar, Dotiwalla, Chouard, Patnaik, Tsui,
	Guo, Zhou, Chai, Caine, et~al.]{sun2019scalability}
	P.~Sun, H.~Kretzschmar, X.~Dotiwalla, A.~Chouard, V.~Patnaik, P.~Tsui, J.~Guo,
	Y.~Zhou, Y.~Chai, B.~Caine, et~al.
	\newblock Scalability in perception for autonomous driving: Waymo open dataset.
	\newblock In \emph{CVPR}, 2020.
	
	\bibitem[Qi et~al.(2018)Qi, Liu, Wu, Su, and Guibas]{qi2018frustum}
	C.~R. Qi, W.~Liu, C.~Wu, H.~Su, and L.~J. Guibas.
	\newblock Frustum pointnets for 3d object detection from rgb-d data.
	\newblock In \emph{CVPR}, pages 918--927, 2018.
	
	\bibitem[Ku et~al.(2018)Ku, Mozifian, Lee, Harakeh, and Waslander]{ku2018joint}
	J.~Ku, M.~Mozifian, J.~Lee, A.~Harakeh, and S.~L. Waslander.
	\newblock Joint 3d proposal generation and object detection from view
	aggregation.
	\newblock In \emph{IROS}. IEEE, 2018.
	
	\bibitem[Wang and Jia(2019)]{wang2019frustum}
	Z.~Wang and K.~Jia.
	\newblock Frustum convnet: Sliding frustums to aggregate local point-wise
	features for amodal 3d object detection.
	\newblock In \emph{IROS}. IEEE, 2019.
	
	\bibitem[Yang et~al.(2018)Yang, Luo, and Urtasun]{yang2018pixor}
	B.~Yang, W.~Luo, and R.~Urtasun.
	\newblock Pixor: Real-time 3d object detection from point clouds.
	\newblock In \emph{CVPR}, 2018.
	
	\bibitem[Simony et~al.(2018)Simony, Milzy, Amendey, and
	Gross]{Simony_2018_ECCV_Workshops}
	M.~Simony, S.~Milzy, K.~Amendey, and H.-M. Gross.
	\newblock Complex-yolo: An euler-region-proposal for real-time 3d object
	detection on point clouds.
	\newblock In \emph{ECCV Workshops}, 2018.
	
	\bibitem[Graham et~al.(2018)Graham, Engelcke, and van~der
	Maaten]{graham2017submanifold}
	B.~Graham, M.~Engelcke, and L.~van~der Maaten.
	\newblock 3d semantic segmentation with submanifold sparse convolutional
	networks.
	\newblock \emph{CVPR}, 2018.
	
	\bibitem[Shi et~al.(2019)Shi, Wang, and Li]{shi2019pointrcnn}
	S.~Shi, X.~Wang, and H.~Li.
	\newblock Pointrcnn: 3d object proposal generation and detection from point
	cloud.
	\newblock In \emph{CVPR}, pages 770--779, 2019.
	
	\bibitem[Shi et~al.(2020)Shi, Guo, Jiang, Wang, Shi, Wang, and Li]{shi2019pv}
	S.~Shi, C.~Guo, L.~Jiang, Z.~Wang, J.~Shi, X.~Wang, and H.~Li.
	\newblock Pv-rcnn: Point-voxel feature set abstraction for 3d object detection.
	\newblock In \emph{CVPR}, pages 10529--10538, 2020.
	
	\bibitem[Yang et~al.(2019)Yang, Sun, Liu, Shen, and Jia]{yang2019std}
	Z.~Yang, Y.~Sun, S.~Liu, X.~Shen, and J.~Jia.
	\newblock Std: Sparse-to-dense 3d object detector for point cloud.
	\newblock In \emph{ICCV}, pages 1951--1960, 2019.
	
	\bibitem[He et~al.(2020)He, Zeng, Huang, Hua, and Zhang]{he2020sassd}
	C.~He, H.~Zeng, J.~Huang, X.-S. Hua, and L.~Zhang.
	\newblock Structure aware single-stage 3d object detection from point cloud.
	\newblock In \emph{CVPR}, 2020.
	
	\bibitem[Jaderberg et~al.(2015)Jaderberg, Simonyan, Zisserman,
	et~al.]{jaderberg2015spatial}
	M.~Jaderberg, K.~Simonyan, A.~Zisserman, et~al.
	\newblock Spatial transformer networks.
	\newblock In \emph{NeurIPS}, pages 2017--2025, 2015.
	
	\bibitem[He et~al.(2017)He, Keuper, Schiele, and Fritz]{he2017learning}
	Y.~He, M.~Keuper, B.~Schiele, and M.~Fritz.
	\newblock Learning dilation factors for semantic segmentation of street scenes.
	\newblock In \emph{German Conference on Pattern Recognition}, pages 41--51.
	Springer, 2017.
	
	\bibitem[Chen et~al.(2019)Chen, Mensink, and Gavves]{chen20193d}
	Y.~Chen, T.~Mensink, and E.~Gavves.
	\newblock 3d neighborhood convolution: Learning depth-aware features for rgb-d
	and rgb semantic segmentation.
	\newblock In \emph{3DV}, pages 173--182. IEEE, 2019.
	
	\bibitem[Wang and Neumann(2018)]{wang2018depth}
	W.~Wang and U.~Neumann.
	\newblock Depth-aware cnn for rgb-d segmentation.
	\newblock In \emph{ECCV}, 2018.
	
	\bibitem[Dai et~al.(2017)Dai, Qi, Xiong, Li, Zhang, Hu, and
	Wei]{dai2017deformable}
	J.~Dai, H.~Qi, Y.~Xiong, Y.~Li, G.~Zhang, H.~Hu, and Y.~Wei.
	\newblock Deformable convolutional networks.
	\newblock In \emph{ICCV}, pages 764--773, 2017.
	
	\bibitem[Strub et~al.(2018)Strub, Seurin, Perez, De~Vries, Mary, Preux, and
	CourvilleOlivier~Pietquin]{strub2018visual}
	F.~Strub, M.~Seurin, E.~Perez, H.~De~Vries, J.~Mary, P.~Preux, and
	A.~CourvilleOlivier~Pietquin.
	\newblock Visual reasoning with multi-hop feature modulation.
	\newblock In \emph{ECCV}, pages 784--800, 2018.
	
	\bibitem[Zhu et~al.(2019)Zhu, Hu, Lin, and Dai]{zhu2019deformable}
	X.~Zhu, H.~Hu, S.~Lin, and J.~Dai.
	\newblock Deformable convnets v2: More deformable, better results.
	\newblock In \emph{CVPR}, pages 9308--9316, 2019.
	
	\bibitem[Ding et~al.(2020)Ding, Huo, Yi, Wang, Shi, Lu, and
	Luo]{ding2020learning}
	M.~Ding, Y.~Huo, H.~Yi, Z.~Wang, J.~Shi, Z.~Lu, and P.~Luo.
	\newblock Learning depth-guided convolutions for monocular 3d object detection.
	\newblock In \emph{CVPR}, 2020.
	
	\bibitem[Chen et~al.(2017)Chen, Papandreou, Kokkinos, Murphy, and
	Yuille]{chen2017deeplab}
	L.-C. Chen, G.~Papandreou, I.~Kokkinos, K.~Murphy, and A.~L. Yuille.
	\newblock Deeplab: Semantic image segmentation with deep convolutional nets,
	atrous convolution, and fully connected crfs.
	\newblock \emph{PAMI}, 40\penalty0 (4), 2017.
	
	\bibitem[Beyer et~al.(2016)Beyer, Hermans, and Leibe]{BeyerHermans2016RAL}
	L.~Beyer, A.~Hermans, and B.~Leibe.
	\newblock {DROW: Real-Time Deep Learning based Wheelchair Detection in 2D Range
		Data}.
	\newblock \emph{RA-L}, 2016.
	
	\bibitem[Yu and Koltun(2016)]{yu2015multi}
	F.~Yu and V.~Koltun.
	\newblock Multi-scale context aggregation by dilated convolutions.
	\newblock In \emph{ICLR}, 2016.
	
	\bibitem[Lin et~al.(2014)Lin, Chen, and Yan]{lin2013network}
	M.~Lin, Q.~Chen, and S.~Yan.
	\newblock Network in network.
	\newblock \emph{ICLR}, 2014.
	
	\bibitem[Goodale et~al.(1992)Goodale, Milner, et~al.]{goodale1992separate}
	M.~A. Goodale, A.~D. Milner, et~al.
	\newblock Separate visual pathways for perception and action.
	\newblock \emph{Trends in Neurosciences}, 15, 1992.
	\newblock ISSN 01662236.
	\newblock \doi{10.1016/0166-2236(92)90344-8}.
	
	\bibitem[Clevert et~al.(2015)Clevert, Unterthiner, and
	Hochreiter]{clevert2015fast}
	D.-A. Clevert, T.~Unterthiner, and S.~Hochreiter.
	\newblock Fast and accurate deep network learning by exponential linear units
	(elus).
	\newblock \emph{arXiv preprint arXiv:1511.07289}, 2015.
	
	\bibitem[Ba et~al.(2016)Ba, Kiros, and Hinton]{ba2016layer}
	J.~L. Ba, J.~R. Kiros, and G.~E. Hinton.
	\newblock Layer normalization.
	\newblock \emph{preprint arXiv:1607.06450}, 2016.
	
	\bibitem[Shi et~al.(2019)Shi, Wang, Wang, and Li]{shi2019part}
	S.~Shi, Z.~Wang, X.~Wang, and H.~Li.
	\newblock Part-a\^{} 2 net: 3d part-aware and aggregation neural network for
	object detection from point cloud.
	\newblock \emph{PAMI}, 2019.
	
	\bibitem[Yu et~al.(2018)Yu, Wang, Shelhamer, and Darrell]{yu2018deep}
	F.~Yu, D.~Wang, E.~Shelhamer, and T.~Darrell.
	\newblock Deep layer aggregation.
	\newblock In \emph{Proceedings of the IEEE conference on computer vision and
		pattern recognition}, pages 2403--2412, 2018.
	
	\bibitem[Lin et~al.(2017)Lin, Goyal, Girshick, He, and
	Doll{\'a}r]{lin2017focal}
	T.-Y. Lin, P.~Goyal, R.~Girshick, K.~He, and P.~Doll{\'a}r.
	\newblock Focal loss for dense object detection.
	\newblock In \emph{ICCV}, pages 2980--2988, 2017.
	
	\bibitem[Ren et~al.(2015)Ren, He, Girshick, and Sun]{ren2015faster}
	S.~Ren, K.~He, R.~Girshick, and J.~Sun.
	\newblock Faster r-cnn: Towards real-time object detection with region proposal
	networks.
	\newblock In \emph{NeurIPS}, 2015.
	
	\bibitem[Geiger et~al.(2013)Geiger, Lenz, Stiller, and
	Urtasun]{geiger2013vision}
	A.~Geiger, P.~Lenz, C.~Stiller, and R.~Urtasun.
	\newblock Vision meets robotics: The kitti dataset.
	\newblock \emph{The International Journal of Robotics Research}, 32\penalty0
	(11):\penalty0 1231--1237, 2013.
	
	\bibitem[Caesar et~al.(2019)Caesar, Bankiti, Lang, Vora, Liong, Xu, Krishnan,
	Pan, Baldan, and Beijbom]{nuscenes2019}
	H.~Caesar, V.~Bankiti, A.~H. Lang, S.~Vora, V.~E. Liong, Q.~Xu, A.~Krishnan,
	Y.~Pan, G.~Baldan, and O.~Beijbom.
	\newblock nuscenes: A multimodal dataset for autonomous driving.
	\newblock \emph{arXiv preprint arXiv:1903.11027}, 2019.
	
	\bibitem[Milioto et~al.(2019)Milioto, Vizzo, Behley, and
	Stachniss]{milioto2019rangenet}
	A.~Milioto, I.~Vizzo, J.~Behley, and C.~Stachniss.
	\newblock Rangenet++: Fast and accurate lidar semantic segmentation.
	\newblock In \emph{IROS}. IEEE, 2019.
	
	\bibitem[Zhou et~al.(2019)Zhou, Sun, Zhang, Anguelov, Gao, Ouyang, Guo, Ngiam,
	and Vasudevan]{zhou2019end}
	Y.~Zhou, P.~Sun, Y.~Zhang, D.~Anguelov, J.~Gao, T.~Ouyang, J.~Guo, J.~Ngiam,
	and V.~Vasudevan.
	\newblock End-to-end multi-view fusion for 3d object detection in lidar point
	clouds.
	\newblock In \emph{CoRL}, 2019.
	
	\bibitem[Kingma and Ba(2015)]{kingma2015adam}
	D.~Kingma and J.~Ba.
	\newblock Adam: A method for stochastic optimization.
	\newblock In \emph{ICLR}, 2015.
	
	\bibitem[Yan et~al.(2018)Yan, Mao, and Li]{yan2018second}
	Y.~Yan, Y.~Mao, and B.~Li.
	\newblock Second: Sparsely embedded convolutional detection.
	\newblock \emph{Sensors}, 2018.
	
	\bibitem[Ngiam et~al.(2019)Ngiam, Caine, Han, Yang, Chai, Sun, Zhou, Yi,
	Alsharif, Nguyen, et~al.]{ngiam2019starnet}
	J.~Ngiam, B.~Caine, W.~Han, B.~Yang, Y.~Chai, P.~Sun, Y.~Zhou, X.~Yi,
	O.~Alsharif, P.~Nguyen, et~al.
	\newblock Starnet: Targeted computation for object detection in point clouds.
	\newblock \emph{arXiv preprint arXiv:1908.11069}, 2019.
	
	\bibitem[Shand(2020)]{shand2020el}
	M.~A. Shand.
	\newblock {Origins and mitigations of some automotive pulsed lidar artifacts}.
	\newblock In B.~Jalali and K.~ichi Kitayama, editors, \emph{AI and Optical Data
		Sciences}, volume 11299, pages 49 -- 55. International Society for Optics and
	Photonics, SPIE, 2020.
	
	\bibitem[He et~al.(2016)He, Zhang, Ren, and Sun]{he2016deep}
	K.~He, X.~Zhang, S.~Ren, and J.~Sun.
	\newblock Deep residual learning for image recognition.
	\newblock In \emph{CVPR}, 2016.
	
	\bibitem[Chen et~al.(2017)Chen, Papandreou, Schroff, and
	Adam]{chen2017rethinking}
	L.-C. Chen, G.~Papandreou, F.~Schroff, and H.~Adam.
	\newblock Rethinking atrous convolution for semantic image segmentation.
	\newblock \emph{arXiv preprint arXiv:1706.05587}, 2017.
	
\end{thebibliography}
\end{document}